\pdfoutput=1
\documentclass[11pt]{article}

\usepackage{authblk}

\usepackage[]{acl}

\usepackage{times}
\usepackage{latexsym}
\usepackage{amsmath}
\usepackage{amssymb}
\usepackage{graphicx}
\usepackage{subcaption}
\usepackage{multirow}
\usepackage{changepage}
\usepackage{hhline}
\usepackage{color}
\usepackage{booktabs}
\usepackage{bbm}
\usepackage{wrapfig,lipsum,booktabs}
\usepackage{makecell}
\usepackage{float}
\usepackage{arydshln}

\usepackage[T1]{fontenc}
\usepackage[utf8]{inputenc}

\usepackage{microtype}

\DeclareMathOperator*{\argmax}{arg\,max}

\title{Semantically Informed Slang Interpretation}

\author[1]{Zhewei Sun}
\author[1,2,4]{Richard Zemel}
\author[1,3,4]{Yang Xu}
\affil[1]{Department of Computer Science, University of Toronto, Toronto, Canada}
\affil[2]{Department of Computer Science, Columbia University, New York, USA}
\affil[3]{Cognitive Science Program, University of Toronto, Toronto, Canada}
\affil[4]{Vector Institute for Artificial Intelligence, Toronto, Canada}
\affil[ ]{\ttfamily \{zheweisun, zemel, yangxu\}@cs.toronto.edu}

\begin{document}
\maketitle
\begin{abstract}
Slang is a predominant form of informal language making  flexible and extended use of words that is notoriously hard for natural language processing systems to interpret. Existing approaches to slang interpretation tend to rely on context but ignore semantic extensions common in slang word usage. We propose a semantically informed slang interpretation (SSI) framework that considers jointly the contextual and semantic appropriateness of a candidate interpretation for a query slang.  We perform rigorous evaluation on two large-scale online slang dictionaries and show that our approach not only achieves state-of-the-art accuracy for slang interpretation in English, but also does so in zero-shot and few-shot scenarios where training data is sparse. Furthermore, we show how the same framework can be applied to enhancing machine translation of slang from English to other languages. Our work creates opportunities for the automated interpretation and translation of informal language.
\end{abstract}

\section{Introduction}
Slang is one of the most common forms of informal language, but interpreting slang can be difficult for both humans and machines. Empirical studies have shown that, although it is done instinctively, interpretation and translation of unfamiliar or novel slang expressions can be quite hard for humans \cite{braun01, mattiello09}. Similarly, slang interpretation is also notoriously difficult for state-of-the-art natural language processing (NLP) systems, which presents a critical challenge to downstream applications such as natural language understanding and machine translation.

%. Central to human cognitive capability of using slang is the ability to both generate and interpret novel slang usages, with varying effectiveness subject to one's social background \cite{labov72, labov06}. 

\begin{figure}[t!]
	%$\begin{center}
	\begin{subfigure}[b]{0.99\linewidth}
		\includegraphics[width=\linewidth]{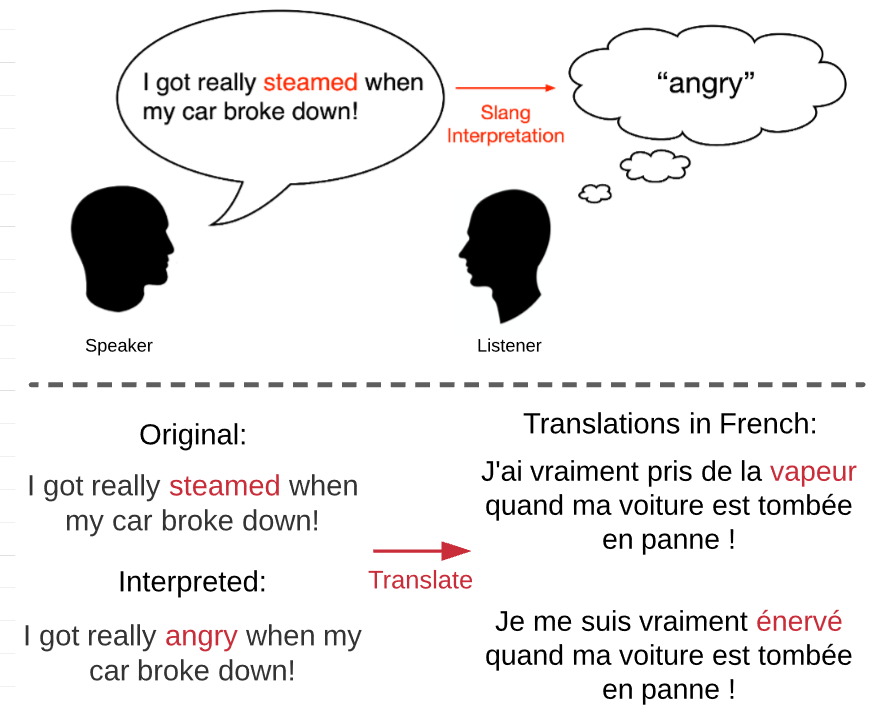}
	\end{subfigure}\vfill
	%\end{center}
	\caption{Illustrations of slang interpretation in English (top panel) and slang translation (bottom panel) from English to French on the original sentence (nonsensical), or on the interpreted version of the sentence (sensical).} 
	\label{fig1}
\end{figure}

Consider the sentence ``I got really \textit{steamed} when my car broke down''. As illustrated in Figure~\ref{fig1}, directly applying a translation system such as Google Translate on this raw English sentence would result in a nonsensical translation of the slang term \textit{steamed} in French. This error is due partly to the underlying language model that fails to recognize the flexible extended use of the slang term from its conventional meaning (e.g., ``vapor'') to the slang meaning of ``angry''. However, if knowledge about such semantic extensions can be incorporated into interpreting the slang prior to translation, as Figure~\ref{fig1} shows the system would be quite effective in translating the intended meaning. 

%whereas the translation system is quite effective once the slang expression has been correctly interpreted.
%

Here we consider the problem of slang interpretation illustrated in the top panel of Figure~\ref{fig1}. Given a target slang term like {\it steamed} in a novel query sentence, we want to automatically infer its intended meaning in the form of a definition (e.g., ``angry''). Tackling this problem has implications in both machine interpretation and understanding of informal language within individual languages and translation between languages.

\begin{figure}[t!]
	%$\begin{center}
	\begin{subfigure}[b]{0.99\linewidth}
		\includegraphics[width=\linewidth]{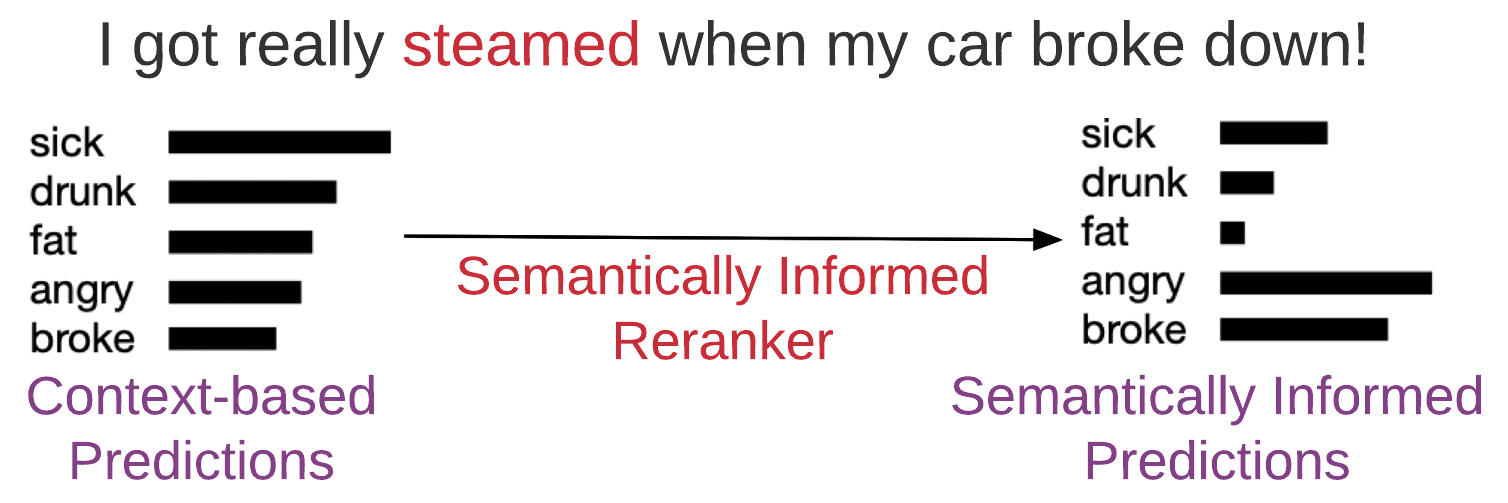}
	\end{subfigure}\vfill
	%\end{center}
	\caption{Workflow of the proposed framework.} 
	\label{fig2}
\end{figure}

One natural solution to this problem is to use contextual information to infer the meaning of a slang term. Figure~\ref{fig2} illustrates this idea by showing the top infilled words predicted under a GPT-2 \citep{radford19} based language infill model \citep{donahue20}. Each of these words can be considered a candidate paraphrase for the target slang {\it steamed} conditioned on its surrounding words.  Although the groundtruth meaning ``angry'' is among the list of top candidates, this model infers ``sick'' as the most probable interpretation. A similar context-based approach has been explored in a previous study led by \citet{ni17} showing that a sequence-to-sequence model trained directly on a large number of pairs of slang-contained sentences along with their corresponding definitions from Urban Dictionary can be a useful starting point toward the automated interpretation of slang.

%pinpoint the intended meaning without considering the conventional meaning of the slang \textit{steamed} (i.e. ``Having been cooked by steaming'').

%However, the usage context is generally under-specified. For example, many possible adjectives could be plausible replacements for the slang expression ``steamed'' in the context sentence ``I got really \_\_\_ when my car broke down''.  

We present an alternative approach to slang interpretation that builds on but goes beyond the context-based models. Inspired by recent work on generative models of slang~\citep{sun19, sun21}, we consider slang interpretation to be the inverse process of slang generation and propose a semantically informed framework that takes into account both contextual information and knowledge about slang meaning extensions (e.g., ``vapor''$\rightarrow$``angry'') in inferring candidate interpretations. Our framework incorporates a semantic model of slang that uses contrastive learning to capture  semantic extensions that link conventional and slang meanings of words~\cite{sun21}. Under this framework, meanings that are otherwise far apart can be brought close, resulting in a semantic space that is sensitive to the flexible extended usages of slang. 
%Rather than using this learned semantic space to generate novel slang usages, we apply it to the inverse problem of slang interpretation by inferring candidate interpretations that are appropriate substitutes for a to-be-interpreted word in slang context.
Rather than using this learned semantic space to generate novel slang usages, we apply it to the inverse problem of slang interpretation by checking whether a candidate interpretation may be suitably expressed as a slang using the to-be-interpreted slang expression.
For example, ``sick'' and ``angry'' can both replace the slang \textit{steamed} in a given context, but ``angry'' may be a more appropriate meaning to be expressed using \textit{steamed} in the slang context. As such, we build a computational framework that takes into account the semantic knowledge of words as well as the context of slang in the interpretation process.

%Inspired by work in psycholinguistics that suggests language generation and comprehension rely on similar cognitive processes~\citep{pickering13, renato20}, we propose  (SSI) framework that considers the generative semantics of slang. 

%Recent work in slang NLP  has shown success in generating novel slang usages by applying contrastive learning on definition entries found in conventional and slang dictionaries. The resulting contrastive sense embeddings (CSE) implicitly encode the extracted semantic extension patterns between corresponding pairs of conventional and slang senses and allow the comparison between sense definitions beyond their surface meaning. 

Figure~\ref{fig2} illustrates the workflow of our approach. We begin with a set of candidate interpretations informed by a context-based model (e.g., a language infill model), where the set would contain a list of possible meanings that fit reasonably in the given context. We then rerank this set of candidate interpretations by selecting the meaning that is most likely to be extended as slang from the to-be-interpreted slang expression.

For the scope of this work, we focus on interpreting slang expressions with existing word forms because extensive studies in slang have suggested that a high proportion of slang usages relies on the extended reuse of existing word forms~\citep{warren92, green10, eble12}. We show that our framework can enhance state-of-the-art language models in slang interpretation in English and slang translation from English to other languages.\footnote{Code and data available at: \url{https://github.com/zhewei-sun/slanginterp}}

%Our work focuses on the semantic aspect of slang instead. That is, the patterns in which senses extend from a slang's original conventional meaning to its intended slang meaning. To facilitate this, we focus on slang expressions with existing word forms that can be found in a standard English dictionary where conventional meaning of the expression can be obtained. 

%The proposed framework can thus be applied to improve any types of generative interpretation model. We show in our experiments that by incorporating semantics into slang interpretation, not only does it improve interpretation accuracy for models trained directly on the slang interpretation task, but also it can enable language models to interpret slang with scarce on-task training data. Furthermore, we experiment with our framework in a machine translation setting, showcasing how translation results can be improved in use cases involving slang.

\section{Related Work}

\subsection{Natural Language Processing for Slang}

Existing approaches in the natural language processing for slang focus on efficient construction, extension, and retrieval from dictionary-based resources for detection~\citep{pal13, dhuliawala16}, interpretation~\citep{gupta19}, and sentiment analysis of slang~\citep{dhuliawala16, wu18}. These studies often rely on heuristic measures to determine or retrieve the meaning of slang and cannot generalize beyond what was available in the training data. Recent work such as \citet{kulkarni18} and \citet{pei19} proposed deep learning based approaches to generalize toward unseen slang.

%Recent work has applied deep learning techniques to slang detection. \citet{pei19} developed an approach that combines linguistic features and a Long-short term memory (LSTM) network~\citep{hochreiter97} with a conditional random field~\citep{lafferty01} and found that while a neural-network based approach is effective, linguistic features such as contextual surprisal are also important to improving slang detection accuracy.

Closely related to our study is \citet{ni17} that formulated English slang interpretation as a translation task (although they did not tackle slang machine translation {\it per se}). In this work, each slang query sentence in English is paired with the groundtruth slang definition (also in English), and such pairs are fed into a translation model. In addition, the spellings of slang word forms are also considered as input. In their model, both the context and the slang form are encoded using separate LSTM encoders. The two encoded representations are then linearly combined to form the encoded input for a sequence-to-sequence network~\citep{sutskever14}. During training, the combined state is passed onto an LSTM decoder to train against the corresponding definition sentence. During test time, beam search \citep{graves12} is applied to decode a set of candidate definition sentences.

One key problem with this approach is that the Dual Encoder tends to rely on the contextual features surrounding the target slang but does not  model flexible meaning extensions of the slang word itself. Similar issues are present in a language-model based approach, whereby one can use an infill model to infer the meaning of a target slang based solely on its surrounding words. Our work extends these context-based approaches by jointly considering the contextual and semantic appropriateness of a slang expression in a sentence, using generative semantic models of slang.

\subsection{Generative Semantic Models of Slang}

%\citet{kulkarni18} developed models to predict the word forms of novel slang using the spelling of its constituent words. In this work, three distinct models were proposed to predict novel blends, clippings, and reduplicatives, all of which are common patterns of slang form extension~\citep{eble12}. 

Recent work by \citet{sun19, sun21} proposed a neural-probabilistic generative framework for modeling slang word choice in novel context. Given a query sentence with the target slang blanked out and the intended meaning of that slang, their framework predicts which word(s) would be appropriate slang choices that fill in the blank. Relevant to their framework is a semantic model of slang that uses contrastive learning from Siamese networks~\citep{Baldi93, Bromley94} to relate conventional and slang meanings of words. This model yields a semantic embedding space that is sensitive to flexible slang meaning extensions. For example, it may learn that meanings associated with ``vapor'' can extend to meanings about ``angry'' (as in the {\it steamed} example in Figure~\ref{fig1}).

%In their framework, definition sentences are first embedded using a sentence embedder such as Sentence-BERT (SBERT; \citealp{reimers19}). Here, few-shot classification is performed by treating each candidate word as a class and its set of conventional meaning as class attributes. The probed slang meaning is then compared to each set of conventional meaning using either 1NN~\citep{koch15, Vinyals16} or prototype~\citep{snell17} similarity of their corresponding SBERT embeddings.

%It has been shown in \citet{sun21} that generation performance can be significantly improved by encoding all senses in a contrastively learned sense embedding space. The contrastive encodings are obtained by training a  with corresponding pairs of conventional and slang meaning as positive learning pairs. The resulting encoding captures semantic extension patterns between corresponding pairs of conventional and slang meaning observed during training. 

Differing from slang generation, our work concerns the inverse problem of slang interpretation that has more direct applications in natural language processing particularly machine translation (e.g., of informal language). Building on work of slang generation, we incorporate the generative semantic model of slang in a semantically informed interpretation framework that integrates context to infer the intended meaning of a target slang.

%\subsection{Reranking}

%In NLP, reranking has been widely applied in both parsing~\citep{charniak05, collins05} and machine translation~\citep{och04, shen04, lee21}. In our slang interpretation framework, we treat a context-based interpreter as our baseline model. We then use knowledge about slang semantic extension to rerank the n-best candidate interpretations generated by the baseline model.

\section{Computational Framework}

Our computational framework is comprised of three key components following the workflow illustrated in Figure~\ref{fig2}: 1) A context-based baseline interpreter that generates an n-best list of candidate interpretations for a target slang in a query sentence; 2) A semantic model of slang that checks the appropriateness of a candidate interpretation to the slang context; 3) A reranker informed by the semantic model in 2) that re-prioritizes the candidate interpretations from the context-based interpreter in 1). We use this framework for both interpreting slang within  English and translating slang from English to other languages.

%We introduce each of the three components in the following sections.

\subsection{Context-based Interpretation} \label{compcontext}

We define slang interpretation formally as follows. Given a target slang term $S$ in context $C_S$ of a query sentence, interpret the  meaning of $S$ by a definition $M$. The context is an important part of the problem formulation since a slang term $S$ may be polysemous and  context can be used to constrain the interpretation of its meaning. We define a slang interpreter $I$ probabilistically as:
\begin{align}
    I(S, C_S) = \argmax_{M} P(M|S, C_S)
\end{align}
Given this formulation, we retrieve an n-best list of candidate interpretations $\mathcal K$ (i.e., $|\mathcal K| = n$) based on an interpretation model of choice $P(M|S, C_S)$. Here, we consider two alternative models for $P(M|S, C_S)$: 1) a language-model (LM) based approach that treats slang interpretation as a cloze task, and 2) a sequence-to-sequence based approach similar to work by \citet{ni17}.

\paragraph{LM-based interpreter.}

The first model we consider is a language infill model in a cloze task, in which the model itself is based on large pre-trained language models such as GPT-2 \citep{radford19}. Although slang expressions may make sporadic appearances during training, this model is not trained specifically on a slang related task and thus serves as a baseline that reflects the state-of-the-art language-model based NLP systems (e.g., \citealp{donahue20}). 

Given context $C_S$ containing target slang $S$, we blank out $S$ in the context and ask the language infill model to infer the most likely words to fill in the blank. This results in a probability distribution $P(w|C_S \backslash S)$ over candidate words $w$. The infilled words can then be viewed as candidate interpretations of the slang $S$:
\begin{align}
    I(S, C_S) = & D[\argmax_{w} LM(w|C_S \backslash S) \nonumber \\
                & + \mathbbm{1}_{T(w)}[T(C_S \backslash S)]] \label{eqlm}
\end{align}
Here, $D$ is a dictionary lookup function that maps a candidate word $w$ to a definition sentence. In this case, we constrain the space of meanings considered to the set of all meanings corresponding to words in the lexicon. Additionally, we apply a Part-of-Speech (POS) tagger $T$ to check whether the candidate word $w$ shares the same POS tag as the blanked-out word in the usage context. Words that share the same POS tags are preferred in the list of n-best retrievals.

This baseline approach by itself does not take into account any (semantic) information from the target slang $S$. In the case where two distinctive slang terms may be placed in the same  context, the model would generate the exact same output. However, this LM based approach does not require task-specific data to train. We show later that by reranking language model outputs, it is possible to achieve state-of-the-art performance using much less on-task data than existing approaches.

\paragraph{Dual encoder.}
\citet{ni17} partly addressed the context-only limitation by encoding the slang term using a character-level recurrent neural network in an end-to-end model inspired by the sequence-to-sequence architecture for neural machine translation \citep{sutskever14}. We implement their dual encoder architecture as an alternative context-based interpreter to LM. In this model, separate LSTM encoders are applied on the context $C_S$ and the character encoding of the to-be-interpreted slang $S$ respectively. The two encoders are then linearly combined using learned parameters.
% \begin{align}
% 	h_{encode} = h_{context}W_{context} + h_{char}W_{char} + B
% \end{align}
The combined state is passed onto an LSTM decoder to train against the corresponding definition sentence in Urban Dictionary (as in the original work of \citealt{ni17}). For inference, beam search \cite{graves12} is applied to decode an n-best list of candidate definition sentences.
% \begin{align}
%     I(S, C_S) = \argmax_{M} Dual(M|S, C_S)
% \end{align}

While this approach is trained directly on slang data and considers the slang word forms, it requires a large on-task dataset to be trained effectively. This model also  does not take into account the appropriateness of meaning extension in slang usage. We next describe how a semantic model of slang can be incorporated to enhance the  context-based interpreters.

\subsection{Semantic Model of Slang} \label{compslang}

Given an n-best list of candidate interpretations $\mathcal K$ for the target slang  $S$ in context $C_S$, we wish to model the semantic plausibility of each candidate interpretation $k\in \mathcal K$. Specifically, we ask how likely one would relate the (conventional meaning of) target slang expression $S$ to a candidate interpretation $k$. \citet{sun19, sun21} modeled the relationship between a to-be-expressed meaning and a word form using the prototype model~\citep{rosch75, snell17}. We adapt this model in the context of slang interpretation:
\begin{align}
    f(k, S) &= sim(E_k, E_S) \nonumber \\
              &= \exp(-\frac{d(E_k, E_S)}{h_{m}})
\label{eqsemantic}
\end{align}
$E_k$ is an embedding for a candidate interpretation $k$ and $E_S$ is the prototypical conventional meaning of $S$ computed by averaging the embeddings of its conventional meanings in dictionary ($\mathcal{E}_S$):
\begin{align}
    E_S = \frac{1}{|\mathcal{E}_S|} \sum_{E_{S_i} \in \mathcal{E}_S} E_{S_i}
\end{align}
The similarity function $f$ can then be computed by taking the negative exponential of the Euclidean distance between the two resulting semantic embeddings. $h_m$ is a kernel width hyperparameter.

Following \citet{sun21}, we learn semantic embeddings $E_k$ and $E_{S_i}$ under a max-margin triplet loss scheme, where embeddings of slang sense definitions ($E_{SL}$) are brought close in Euclidean space to those of their conventional sense definitions ($E_P$) yet kept apart from irrelevant word senses ($E_N$) by a pre-specified margin $m$:
% 	\begin{adjustwidth}{-0.2cm}{}
\begin{align}
	Loss &= \Big[ d(E_{SL}, E_P) - d(E_{SL}, E_N) + m \Big]_+ \label{eqmmt}
\end{align}
% 	\end{adjustwidth}

The resulting contrasive sense encodings are shown to be sensitive to slang semantic extensions that have been observed during training. We leverage this knowledge to check whether pairing a candidate interpretation $k$ with the slang expression $S$ is likely given the common semantic extensions observed in slang usages.% The resulting scores can then be used to rerank the candidate interpretations.

\subsection{Semantically Informed Reranking}

We define a semantic scorer $g$ over the set of candidate interpretations $\mathcal K$ and the to-be-interpreted slang $S$. The candidates are reranked based on the resulting scores to obtain semantically informed slang interpretations (SSI):
\begin{align}
	SSI(\mathcal K) = \argmax g(k,S)
\end{align}
We define $g(\mathcal K,S)$ as a score distribution over the set of candidates $\mathcal K$ given slang $S$, where each score is computed by checking the semantic appropriateness of a candidate meaning $k\in \mathcal K$ with respect to target slang  $S$ by querying the semantic model $f$ from Equation~\ref{eqsemantic}:
\begin{align}
	g(k,S) = P(k|S) \propto f(k, S)
\end{align}

In addition, we apply collaborative filtering \citep{goldberg92} to account for a small neighborhood of words $L(S)$ akin to the slang expression $S$ in conventional meaning:
\begin{align}
	g^*(k,S) \propto \sum_{S' \in L(S)} sim(S, S')g(k,S') \label{eqls}\\
	sim(S, S') = \exp(-\frac{d(S, S')}{h_{cf}}) \label{eqftsim}
\end{align}

Here, $d(S, S')$ is the cosine distance between the two slang's word vectors and $h_{cf}$ is a hyperparameter controlling the kernel width. The collaborative filtering step encodes intuition from studies in historic semantic change that similar words tend to extend to express similar meanings \citep{lehrer85, xu15}, which was found to extend well in the case of slang \citep{sun19,sun21}.

\section{Datasets}

\begin{table*}[t!]
	\centering\makebox[\textwidth]{
		\begin{tabular}{lccccccc}
			Dataset&\makecell{\# of unique \\slang word forms}&\makecell{\# of slang \\definition entries}&\makecell{\# of context \\sentences}&\makecell{\# of definitions\\in the test set}&\makecell{\# of context sentences\\in the test set}
			\\
			\addlinespace[0.05cm]
			\hline
			\addlinespace[0.15cm]
			OSD&1,635&2,979&3,718&299&405\\
			\addlinespace[0.15cm]
			UD&9,474&65,478&65,478&1,242&1,242\\
			\addlinespace[0.15cm]
	\end{tabular}}
	\caption{Summary of basic statistics for the two online slang dictionaries used in the study.}
	\label{tabledata}
\end{table*}

We use two online English slang dictionary resources to train and evaluate our proposed slang interpretation framework: 1) the Online Slang Dictionary (OSD)\footnote{OSD: http://onlineslangdictionary.com} dataset from \citet{sun21} and 2) a collection of Urban Dictionary (UD)\footnote{UD: https://www.urbandictionary.com} entries from 1999 to 2014 collected by \citet{ni17}. Each dataset contains slang gloss entries including a slang's word form, its definition, and at least one corresponding example sentence containing the slang term. We use the same training and testing split provided by the original authors and only use entries where a corresponding non-informal entry can be found in the online version of the Oxford Dictionary (OD) for English\footnote{OD: https://en.oxforddictionaries.com}, which allows the retrieval of conventional senses for all slang expressions considered. We also filter out entries where the example usage sentence contains none or more than one exact references of the corresponding slang expression. When a definition entry has multiple example usage sentences, we treat each example sentence as a separate data entry, but all data entries corresponding to the same definition entry will only appear in the same data split. Table~\ref{tabledata} shows the size of the datasets after pre-processing. While OSD contains higher quality entries, UD offers a much larger dataset. We thus use OSD to evaluate model performance in a low resource scenario and UD for evaluation of larger neural network based approaches.

\section{Evaluation and Results}

\subsection{Evaluation on Slang Interpretation} \label{evalmodel}

\begin{table*}[t!]
	\centering\makebox[\textwidth]{
		\begin{tabular}{lrr}
			Model&\makecell[r]{Distinctively\\sampled candidates}&\makecell[r]{Randomly\\sampled candidates}
			\\
			\addlinespace[0.05cm]
			\hline
			%\addlinespace[0.1cm]
			\addlinespace[0.05cm]
			Dataset 1: Online Slang Dictionary (OSD) \citep{sun21}\\
			%\addlinespace[0.25cm]
			\addlinespace[0.1cm]
			Language Infill Model (LM Infill) \citep{donahue20}, $n$ = 50&0.532&0.502\\
			%\addlinespace[0.1cm]
			\hspace{0.5cm} + Semantically Informed Slang Interpretation (SSI)&\textbf{0.557}&\textbf{0.563}\\
			\hdashline
			%\addlinespace[0.2cm]
			\addlinespace[0.1cm]
			Dual Encoder*~\citep{ni17}, $n$ = 5&0.584&0.583\\
			%\addlinespace[0.1cm]
			\hspace{0.5cm} + SSI &\textbf{0.592}&\textbf{0.588}\\
			%\addlinespace[0.1cm]
			Dual Encoder*, $n$ = 50&0.568&0.602\\
			%\addlinespace[0.1cm]
			\hspace{0.5cm} + SSI&\textbf{0.616}&\textbf{0.607}\\
			\small * Dual Encoders trained on UD data after filtering out slang in OSD test set.&&\\
			%\addlinespace[0.15cm]
			\addlinespace[0.05cm]
			\hline
            %\addlinespace[0.1cm]
            \addlinespace[0.05cm]
			Dataset 2: Urban Dictionary (UD) \citep{ni17}\\
			%\addlinespace[0.25cm]
			\addlinespace[0.1cm]
			LM Infill, $n$ = 50&0.517&0.521\\
			%\addlinespace[0.1cm]
			\hspace{0.5cm} + SSI&\textbf{0.569}&\textbf{0.579}\\
			\hdashline
			%\addlinespace[0.2cm]
			\addlinespace[0.1cm]
			Dual Encoder, $n$ = 5&0.556&0.555\\
			%\addlinespace[0.1cm]
			\hspace{0.5cm} + SSI &\textbf{0.573}&\textbf{0.572}\\
			%\addlinespace[0.1cm]
			Dual Encoder, $n$ = 50&0.547&0.550\\
			%\addlinespace[0.1cm]
			\hspace{0.5cm} + SSI&\textbf{0.582}&\textbf{0.584}\\
	\end{tabular}}
	\caption{Evaluation of English slang interpretation measured in mean-reciprocal rank (MRR). Predictions are ranked against 4 negative candidates  distinctively or randomly sampled, yielding MRR$=$0.457 for the random baseline.}
	\label{tableresults}
\end{table*}

We first evaluate the semantically informed and baseline interpretation models in a multiple choice task. In this task, each query is paired with a set of definitions that construe the meaning of the target slang in the query. One of these definitions is the groundtruth meaning of the target slang, while the other definitions are incorrect or negative entries sampled from the training set (i.e., all taken from the slang dictionary resources described). To score a model, each definition sentence is first compared with the model-predicted definition by computing the Euclidean distance between their respective Sentence-BERT \cite{reimers19} embeddings. The ideal model should produce a definition that is semantically closer to the groundtruth definition, more so than the other competing negatives. For each dataset, we sample two sets of negatives. The first set of negative candidates contains only definition sentences from the training set that are distinct from the groundtruth definition. We consider two definition sentences to be distinct if the overlap in the number of content words is less than 50\%. The other set of negative definitions is sampled randomly. We measure the performance of the models by computing the standard mean reciprocal rank (MRR) of the groundtruth definition's rank when checked against 4 other sampled negative definitions.

We train the semantic reranker on all definition entries in the respective training sets from the two data resources. When training the Dual Encoder, we use 400,431 out-of-vocabulary slang entries (i.e., entries with a slang expression that does not contain a corresponding lexical entry in the standard dictionary) from UD in addition to the in-vocabulary entries used to train the reranker. This is necessary since the baseline Dual Encoder performs poorly without a large number of training entries. Similarly, training the Dual Encoder directly on the OSD training set does not result in an adequate model for comparison. We instead train the Dual Encoder on all UD entries and experiment with the resulting interpreter on OSD. Any UD entries corresponding to words found in the OSD testset are filtered out in this particular experiment. Detailed training procedures for all models can be found in Appendix~\ref{apptrain}.

\begin{table*}[t!]
	\centering\makebox[\textwidth]{
		\begin{tabular}{ll}
			\hline
			
			% Example 1
			\addlinespace[0.1cm]
			\addlinespace[0.05cm]
			%[Example 1]&\\
			\addlinespace[0.1cm]
			Query (target slang in \textit{\textbf{bold italic}}): &That chick is \textit{\textbf{lit}}!\\
			%\addlinespace[0.1cm]
			Groundtruth definition of target slang: &Attractive.\\ \hdashline
			%\addlinespace[0.25cm]
			\addlinespace[0.1cm]
			LM Infill baseline prediction: &Cute, beautiful, adorable.\\
			%\addlinespace[0.1cm]
			LM Infill + SSI prediction: &Hot, cool, fat.\\
			%\addlinespace[0.25cm]
			\addlinespace[0.1cm]
			Dual Encoder baseline prediction: &Another word for bitch.\\
			%\addlinespace[0.1cm]
			Dual Encoder + SSI prediction: &Word used to describe someone who is very attractive.\\
			%\addlinespace[0.15cm]
			\addlinespace[0.05cm]
           \hline
            
            % Example 7
			%\addlinespace[0.1cm]
			\addlinespace[0.05cm]
			%[Example 7]&\\
			%\addlinespace[0.1cm]
			Query: &That Louis Vuitton purse is \textit{\textbf{lush}}!\\
			%\addlinespace[0.1cm]
			Groundtruth definition of target slang: &High quality, luxurious.  (British slang.)\\
			%\addlinespace[0.25cm]
			\hdashline
			\addlinespace[0.1cm]
			LM Infill baseline prediction: &Amazing, beautiful, unique.\\
			%\addlinespace[0.1cm]
			LM Infill + SSI prediction: &Lovely, stunning, expensive.\\
			%\addlinespace[0.25cm]
			\addlinespace[0.1cm]
			Dual Encoder baseline prediction: &Something that is cool or awesome.\\
			%\addlinespace[0.1cm]
			Dual Encoder + SSI prediction: &An adjective used to describe something that is not cool.\\
			%\addlinespace[0.15cm]
		%	\addlinespace[0.05cm]
        %    \hline

	    \end{tabular}
	}
	\caption{Example queries from OSD and top predictions made from both the baseline language infill models (LM Infill) and the Dual Encoder models with $n$ = 50, along with top predictions from the enhanced semantically informed slang interpretation (SSI) models. Additional examples can be found in Appendix~\ref{appex}.
}
	\label{tableexamplemain}
\end{table*}

Table~\ref{tableresults} summarizes the multiple-choice evaluation results on both slang datasets. In all cases, applying the semantically informed slang interpretation framework improves the MRR of the respective baselines under both types of negative candidate sampling. On the UD evaluation, even though the language infill model (LM Infill) is not trained on this specific task, LM infill based SSI is able to select better and more appropriate interpretations than the dual encoder baseline, which is trained specifically on slang interpretation with more than 7 times the number of definition entries for training. We also find that while increasing the beam size (specified by $n$) in the sequence-to-sequence based Dual Encoder model impairs its performance, SSI can take advantage of the additional variation in the generated candidates and outperform its counterpart with a smaller beam size.

Table~\ref{tableexamplemain} provides example interpretations predicted by the models. The \textit{lit} example shows a case where the semantically informed models were able to correctly pinpoint the intended definition, among alternative definitions that describe individuals. The \textit{lush} example suggests that the SSI model is not perfect and points to common errors made by the model including predicting definitions that are more general and applying incorrect semantic extensions. In this case, the model predicts the slang \textit{lush} to mean ``something that is not cool'' because polarity shift is a common pattern in slang usage~\cite{eble12}, even though the groundtruth definition does not make such a polarity shift in this specific example.

Note that the improvement brought by SSI is less prominent in the OSD experiment where the Dual Encoder trained on UD was used. This is expected because the Dual Encoder is trained to generate definition sentences in the style of UD entries, whereas the SSI is trained on OSD definition sentences instead. The mismatch in style between the two datasets might have caused the difference in performance gain.

%We also examine the affect of context length in the usage sentences to interpretation performance on both datasets. We find no significant trend in performance for both the baseline models and the semantically informed variants. Detailed results can be found in Appendix~\ref{appcontext}.

\subsection{Zero-shot and Few-shot Interpretation}

Recent studies in deep learning have shown that large neural network based models such as GPT-3 excel at learning new tasks in a few-shot learning setting~\cite{brown20}. We examine to what extent the superior performance of our SSI framework may be affected by fine-tuning the LM baseline model in zero-shot and few-shot scenarios. We finetune the language infill model (LM Infill) on the first example usage sentence that correspond to each definition entry in the OSD dataset, resulting in 2,979 sentences. Given an example  sentence, we mask out the slang expression and train the language infill model to predict the corresponding slang term. We randomly shuffle all examples and finetune LM Infill for one epoch. We then compare the resulting model with the off-the-shelf LM using examples in the test set that were not used in finetuning (i.e., entries with usage sentences that do not correspond to the first example usage sentence of a definition entry). This results in 106 novel examples for evaluation.

Table~\ref{tablefew} shows the result of this experiment. While finetuning does improve test performance (a 6 point gain in MRR), it remains beneficial to consider semantic information in slang context. In both the zero-shot and the few-shot cases, SSI brings significant performance gain even though SSI itself is only trained on entries from the training set.

\begin{table}[t!]
	\centering\makebox[0.5\textwidth]{
		\begin{tabular}{lrr}
			Model&\makecell[r]{Distinct  \\negatives}&\makecell[r]{Random  \\negatives}
			\\
			\addlinespace[0.05cm]
			\hline
			%\addlinespace[0.25cm]
			\addlinespace[0.1cm]
			LM Zero-shot, $n$ = 50 &0.444&0.443\\
			%\addlinespace[0.1cm]
			\hspace{0.5cm} + SSI&\textbf{0.571}&\textbf{0.565}\\
			%\addlinespace[0.2cm]
			\addlinespace[0.1cm]
			LM Few-shot, $n$ = 50 &0.504&0.513\\
			%\addlinespace[0.1cm]
			\hspace{0.5cm} + SSI &\textbf{0.567}&\textbf{0.564}\\
	\end{tabular}}
	\caption{Interpretation results on OSD measured in mean-reciprocal rank (MRR) before and after finetuning the language infill model.}
	\label{tablefew}
\end{table}

\subsection{Evaluation on Slang Translation} \label{restrans}

We next apply the slang interpretation framework to neural machine translation. Existing machine translation systems have difficulty in translating source sentences containing slang usage partly because they lack the ability to properly decode the  intended slang meaning. We make a first attempt in addressing this problem by exploring whether machine interpretation of slang can lead to better translation of slang. Given a source English sentence containing a slang expression $S$, we apply the LM based slang interpreters to generate a paraphrased word to replace $S$. The paraphrased sentence would then contain the intended meaning of the slang in its literal form. Here, we take advantage of the LM-based approaches' ability to directly generate a paraphrase instead of a definition sentence (i.e., without dictionary lookup $D$ in Equation~\ref{eqlm}), which allows direct insertion of the resulting interpretation into the original sentence.

\begin{figure*}[t!]
    \begin{subfigure}[b]{0.49\linewidth}
		\includegraphics[width=\linewidth]{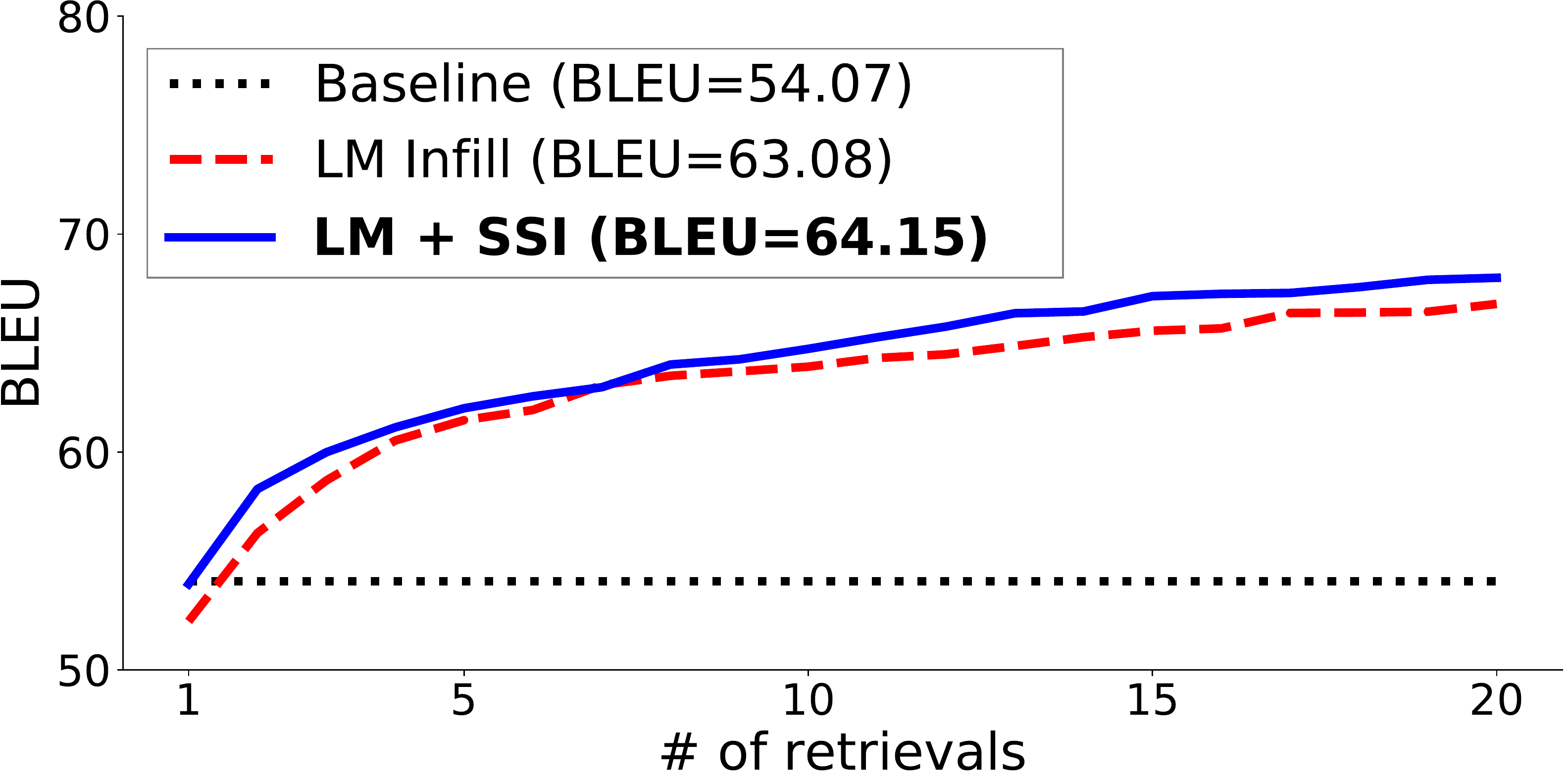}
		%\caption{English to French}
	\end{subfigure}\
	\begin{subfigure}[b]{0.49\linewidth}
		%\vspace{0.2cm}
		\includegraphics[width=\linewidth]{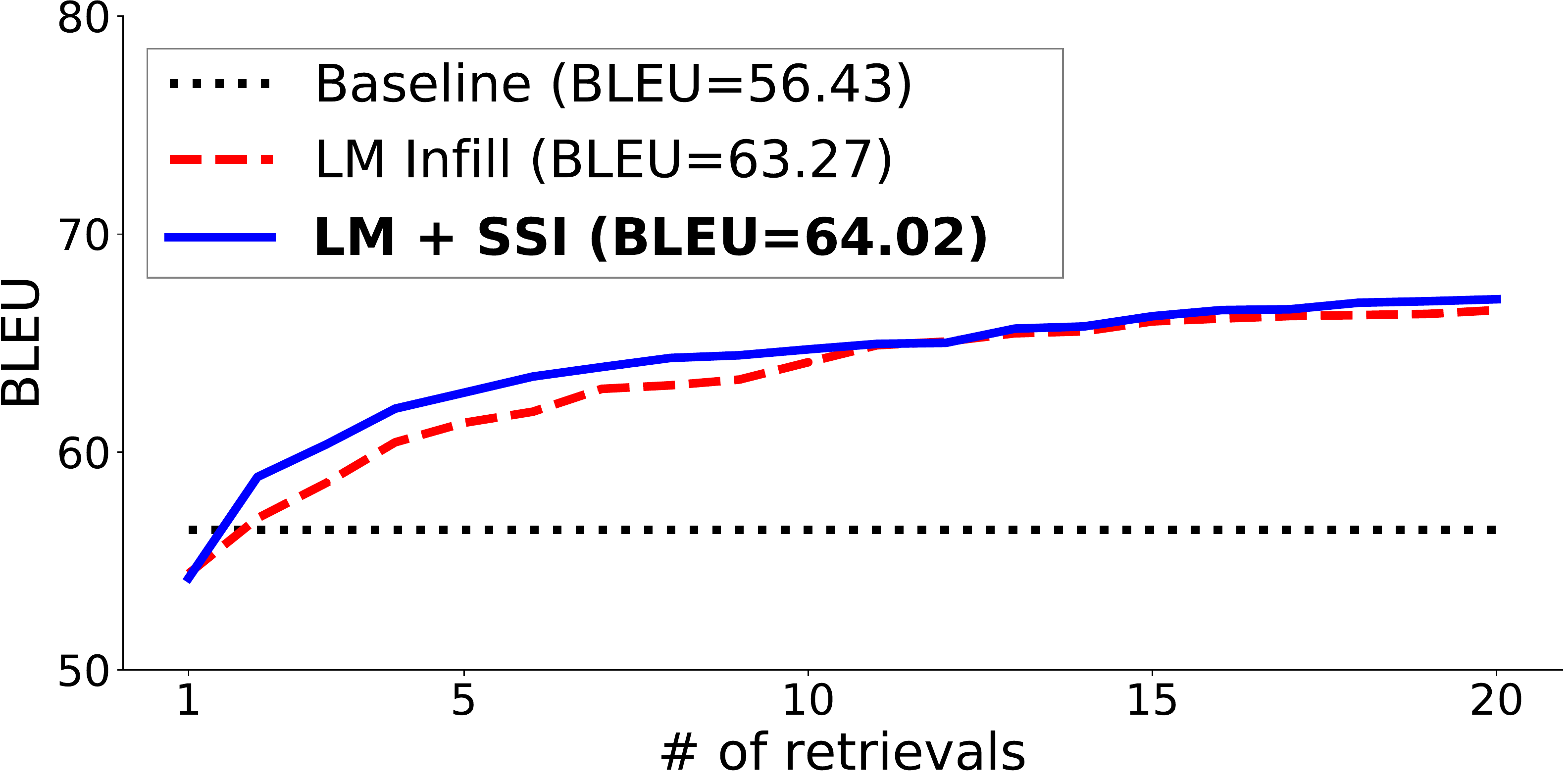}
		%\caption{English to German}
	\end{subfigure}
	\begin{subfigure}[b]{0.49\linewidth}
		\includegraphics[width=\linewidth]{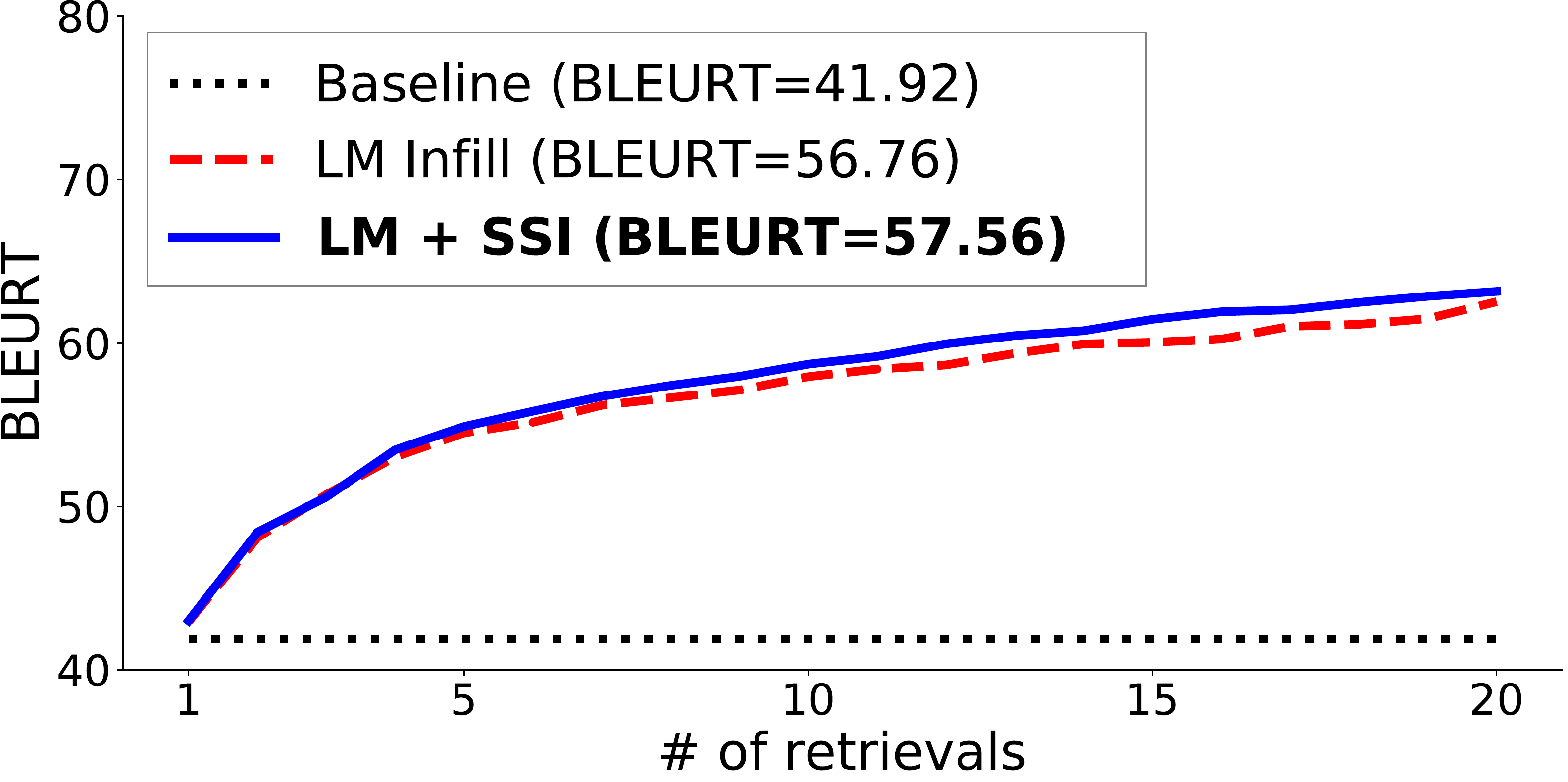}
		%\caption{English to French}
	\end{subfigure}\
	\begin{subfigure}[b]{0.49\linewidth}
		%\vspace{0.2cm}
		\includegraphics[width=\linewidth]{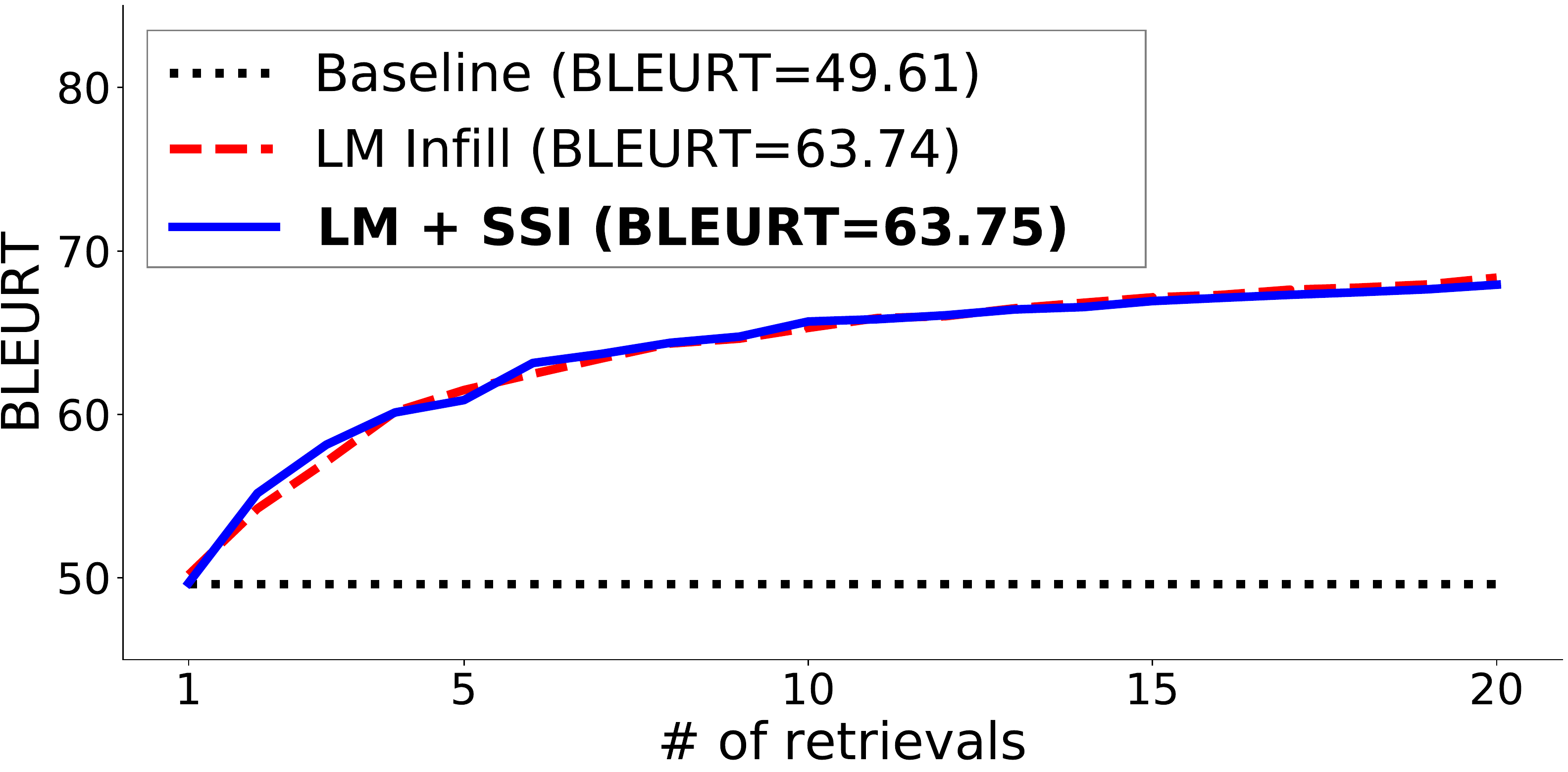}
		%\caption{English to German}
	\end{subfigure}
	\begin{subfigure}[b]{0.49\linewidth}
		\includegraphics[width=\linewidth]{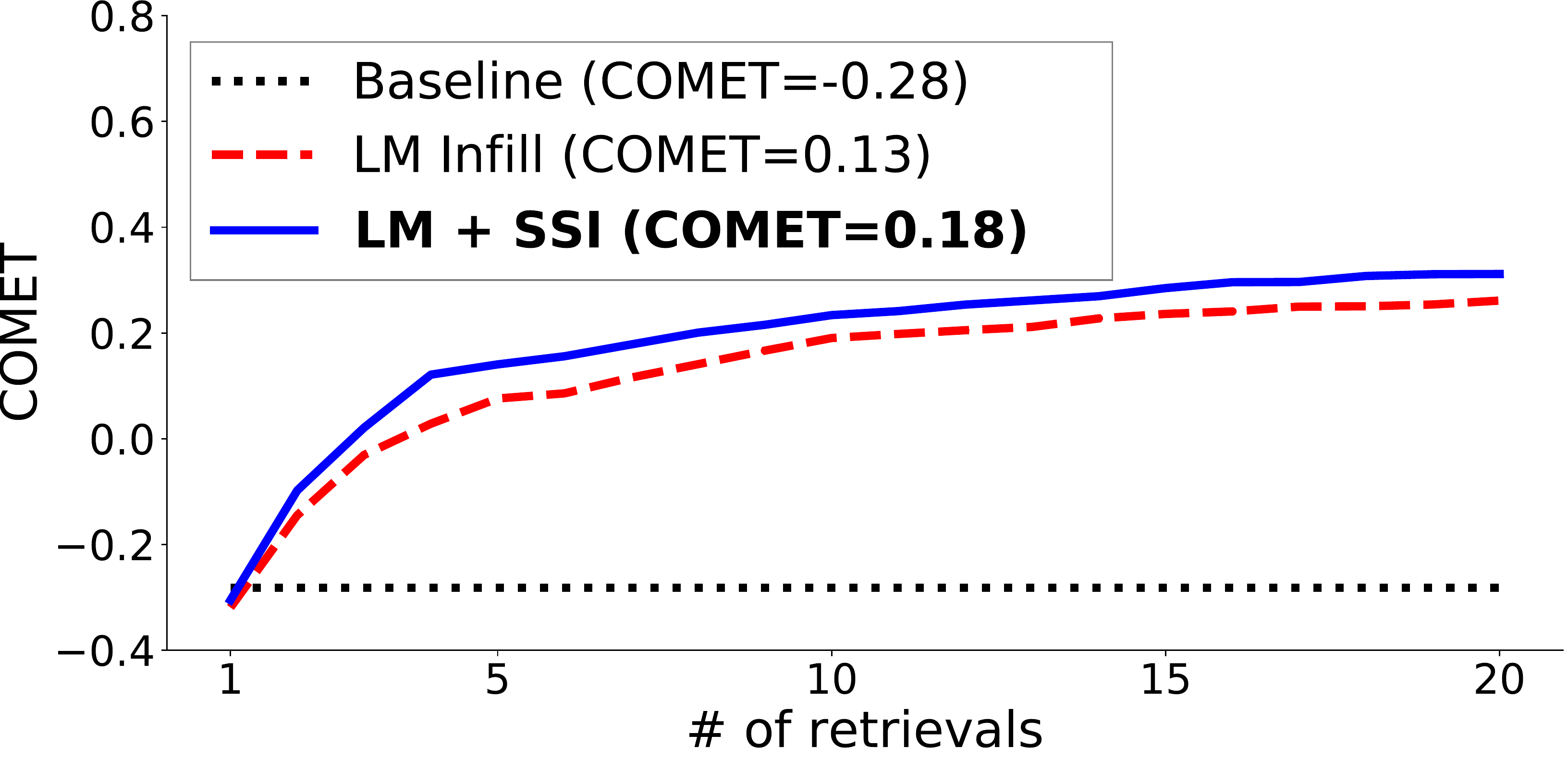}
		\caption{English to French}
	\end{subfigure}\
	\hspace{0.1cm}
	\begin{subfigure}[b]{0.49\linewidth}
		%\vspace{0.2cm}
		\includegraphics[width=\linewidth]{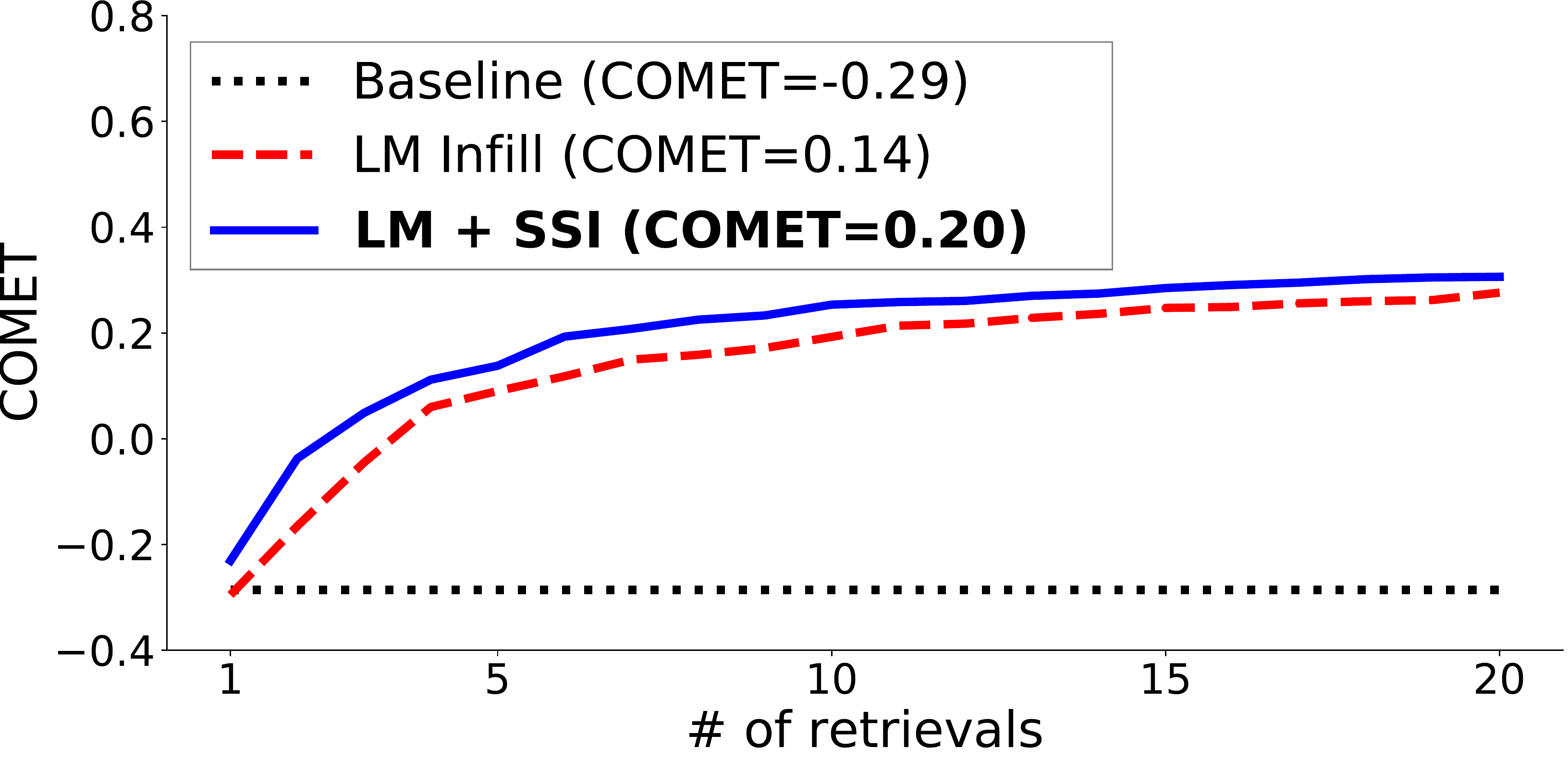}
		\caption{English to German}
	\end{subfigure}
	\caption{Translation scores of translated sentences with the slang replaced by n-best interpretations. Curves show sentence-level BLEU, BLEURT, and COMET scores of the best translation within the top-n retrievals. Aggregate scores integrated over the first 20 retrievals are shown in parenthesis. Baselines are obtained by directly translating the original sentence containing slang.} 
	\label{transfig}
\end{figure*}

%We perform our experiment on the OSD test set because it provides both the groundtruth definitions of slang and the corresponding example sentences which are critical to evaluating the quality of slang translation.
We perform our experiment on the OSD test set because it contains higher quality example sentences than UD.
To mitigate potential biases, we consider only entries that correspond to single word slang expressions, and that the slang has not been seen during training (where the slang attaches to a different slang meaning than the one in the test set). For the remaining 102 test entries, we obtain gold-standard translations by first manually replacing the slang word in the example sentence with its intended definition, condensed to a word or short phrase to fit into the context sentence. We then translate the sentences to French and German using machine translation.%\footnote{Data available for review at: {placeholder URL}}

We make all machine translations using pre-trained 6-layer transformer networks~\citep{vaswani17} from MarianMT~\citep{tiedemann20}, which are trained on a collection of web-based texts in the OPUS dataset~\citep{tiedemann12}. Here, we select models pre-trained on web-based texts to maximize the baseline model's ability to correctly process slang. We evaluate the translated sentences using three metrics: 1) Sentence-level BLEU scores~\citep{papineni02} computed using  \textit{sentence\_bleu} implementation from NLTK~\citep{bird2009natural} with smoothing (\textit{method4} in NLTK, \citealp{chen14}) to account for sparse n-gram overlaps; 2) BLEURT scores~\cite{sellam20} computed using the pre-trained \textit{BLEURT-20} checkpoint; 3) COMET scores~\cite{rei20} computed using the pre-trained \textit{wmt20-comet-da} checkpoint. For COMET scores, we replace slang expressions in the source sentences with their literal equivalents to reduce confusion that the COMET model might have on slang.

\begin{table*}[t!]
	\centering\makebox[\textwidth]{
		\begin{tabular}{ll}
			\hline
            
            % Example 4
			%\addlinespace[0.1cm]
			%[Example 4]&\\
			%\addlinespace[0.1cm]
			Query (target slang in \textit{\textbf{bold italic}}): &I want to go get coffee but it's \textit{\textbf{bitter}} outside.\\
			Definition of target slang: &Abbreviated form of bitterly cold.\\
			Groundtruth interpreted sentence: &I want to go get coffee but it's \textit{bitterly cold} outside.\\
			%\addlinespace[0.25cm]
			\addlinespace[0.1cm]
			 Original query sentence translation: &Je veux aller prendre un café mais c'est amer dehors.\\
			 &\small(\textit{BLEU:} 65.0, \textit{BLEURT:} 59.8, \textit{COMET:} 0.77)\\
			Gold-standard  translation: &Je veux aller prendre un café, mais il fait très froid dehors.\\
			%\addlinespace[0.25cm]
		\hdashline	\addlinespace[0.1cm]
			LM Infill interpretation \& translation: &\\
			%\addlinespace[0.1cm]
			(1) I want to go get coffee but it's \textit{raining} outside. & Je veux aller prendre un café mais il \textit{pleut} dehors.\\
			&\small(\textit{BLEU:} 68.1, \textit{BLEURT:} 79.9, \textit{COMET:} 0.97)\\
            (2) I want to go get coffee but it's \textit{closed} outside. & Je veux aller prendre un café mais il \textit{est fermé} dehors.\\
            &\small(\textit{BLEU:} 70.7, \textit{BLEURT:} 53.9, \textit{COMET:} -0.15)\\
            %(3) I want to go get coffee but it's pouring outside.& Je veux aller chercher du café, mais ça coule dehors. (51.9)\\
            %(4) I want to go get coffee but it's been outside. & Je veux aller prendre un café, mais ça a été dehors. (68.4)\\
            %(5) I want to go get coffee but it's starting outside & Je veux aller prendre un café, mais ça commence dehors. (68.5)\\
            %\addlinespace[0.25cm]
            \addlinespace[0.1cm]
			LM Infill + SSI interpretation \& translation: &\\
			%\addlinespace[0.1cm]
			(1) I want to go get coffee but it's \textit{cold} outside. & Je veux aller prendre un café, mais il fait \textit{froid} dehors.\\
			&\small(\textit{BLEU:} 90.3, \textit{BLEURT:} 92.7, \textit{COMET:} 1.20)\\
            (2) I want to go get coffee but it's \textit{warm} outside. & Je veux aller prendre un café mais il fait \textit{chaud} dehors.\\
            &\small(\textit{BLEU:} 78.1, \textit{BLEURT:} 79.1, \textit{COMET:} 1.12)\\
            %(3) I want to go get coffee but it's driving outside. & Je veux aller prendre un café mais il conduit dehors. (70.4)\\
            %(4) I want to go get coffee but it's closing outside. & Je veux aller prendre un café mais il se ferme dehors. (69.8)\\
            %(5) I want to go get coffee but it's dark outside. & Je veux aller prendre un café, mais il fait noir dehors. (82.3)\\

			%\addlinespace[0.15cm]
			%\addlinespace[0.05cm]

	\end{tabular}}
	\caption{An example of machine translation of slang, without or with the application of the SSI framework. The top 2 interpreted and translated sentences are shown for each model with BLEU, BLEURT, and COMET scores against the gold-standard translation shown in parentheses. More examples can be found in Appendix~\ref{apptrans}.}
	\label{tableextransmain}
\end{table*}

%\footnote{We used the \textit{sentence\_bleu} implementation from NLTK~\citep{bird2009natural} with method4 smoothing.}.

Figure~\ref{transfig} summarizes the results. Overall, the semantically informed approach tends to outperform the baseline approaches for the range of top retrievals (from 1 to 20) under all three metrics considered, with the exception of BLEURT evaluated on German where the semantically informed approach gives very similar performance as the language model baseline. While not all predicted interpretations correspond to the groundtruth definitions, the set of interpreted sentences often contain plausible interpretations that result in improved translation of slang. Table~\ref{tableextransmain} provides some example translations. We observe that quality translations can be found reliably with a small number of interpretation retrievals (i.e., around 5) and the quality generally improves as we retrieve more candidate interpretations. Our approach may be ultimately integrated with a slang detector (e.g., \citealt{pei19}) to produce fully automated translations in natural context that involves slang.

%Future work in machine translation can potentially benefit from the extra variety of translation candidates and improve translations for use cases involving slang.

\section{Conclusion}

The flexible nature of slang is a hallmark of informal language, and to our knowledge we have presented the first principled framework for automated slang interpretation that takes into account both contextual information and knowledge about semantic extensions of slang usage. We showed that our framework is more effective in interpreting and translating the meanings of English slang terms in natural sentences in comparison to existing approaches that rely more heavily on context to infer slang meaning.

Future work in this area may benefit from principled approaches that model the coinage of slang expressions with novel word forms and multi-word  expressions with complex formation strategies, as well as how slang terms emerge in specific individuals and groups. Our current study shows promise for advancing methodologies in informal language processing toward these avenues of future research.

%Our study shows promise for advancing methodologies on informal language processing, and it also points to future avenues of research in this area such as the understanding of complex expressions and novel word forms in informal language use. 

%To our knowledge, this study offers the  the first principled approach to explicitly model slang semantics in slang interpretation and our results show that interpretations can indeed be improved by considering their semantic appropriateness. Our framework will provide opportunities for future research in important downstream tasks such as machine translation of slang.

\section*{Ethical Considerations}

We analyze entries of slang usage in our work and acknowledge that such usages may contain offensive information. We retain such entries in our datasets to preserve the scientific validity of our results, as a significant portion of slang usage aligns to possibly offensive usage context. In the presentation our of results, however, we strive to select examples or illustrations that minimize the extent to which offensive content is represented. We also acknowledge that models trained on datasets such as the Urban Dictionary have a greater tendency to generate offensive language. All model outputs shown are results of model learning and do not reflect opinions of the authors and their affiliated organizations. We hope that our work will contribute to the greater good by enhancing AI system's ability to comprehend such offensive language use, allowing better filtering of online content that may be potentially harmful.

%We thank Graeme Hirst, Suzanne Stevenson, and Lei Yu for offering thoughtful feedback on this work. and Derek Denis for stimulating discussion

\section*{Acknowledgements}
	
We thank the  ARR reviewers for their constructive comments and suggestions, and Walter Rader for  permission to use The Online Slang Dictionary. This work was supported by a NSERC Discovery
Grant RGPIN-2018-05872, a SSHRC Insight Grant
\#435190272, and an Ontario ERA Award to YX.

\bibliography{acl_latex}
\bibliographystyle{acl_natbib}

\clearpage

\appendix

\section{Training Procedures} \label{apptrain}

\subsection{Baseline Models}

We train two context-based slang interpreters described in Section~\ref{compcontext} as our baseline models. For the LM-based interpreter, we use a pre-trained language infill model from \citet{donahue20} based on the GPT-2~\citep{radford19} architecture. Here, we obtain the n-best list of interpretations by retrieving the list of infilled words with the highest infill probability. Words containing non-alphanumeric characters are filtered out. For the dictionary lookup function $D$ in Equation~\ref{eqlm}, if a matching dictionary entry can be found in Oxford Dictionary (OD), the top definition sentence is retrieved as the definition sentence for the input word. Otherwise, the word itself is used as the definition. In addition to the word‘s original form, we apply lemmatization or stemming to the original form using NLTK~\citep{bird2009natural} to find matching dictionary entries. To check for Part-of-Speech (POS) tags, we apply the Flair tagger~\citep{akbik18} on the context sentence with the slang expression replaced by a mask token and use counts from Histwords~\citep{hamilton16} to determine POS tags for individual words.

To train the Dual Encoder, we use LSTM encoders with 256 and 1024 hidden units to encode a slang expression's spelling and its usage context respectively, with 100 and 300 dimensional input embeddings for the characters and words respectively. Following \citet{ni17}, we use random initialization for the input embeddings and use stochastic gradient descent (SGD) with an adaptive learning rate. We train the model for 20 epochs beginning with a learning rate of 0.1 and add an exponential decay of 0.9 every epoch. We reserve 5\% of the training examples as a development set for hyperparameter tuning. We train the model for 20 epochs on a Nvidia Titan V GPU and took 12 hours to complete. During inference, we obtain the n-best list of interpretations by running a beam search of corresponding beam width on the LSTM decoder.

\subsection{Semantic Reranker}

We obtain the contrastive sense encodings (CSE) described in Section~\ref{compslang} by using 768-dimensional Sentence-BERT~\citep{reimers19} embeddings as our baseline embedding. Following \citet{sun21}, we train the contrastive network with a 1.0 margin ($m$ in Equation~\ref{eqmmt}) using Adam~\cite{kingma15} with a learning rate of $2^{-5}$, resulting in 768-dimensional definition sense presentations. We reserve 5\% of the training examples as a development set for hyperparameter tuning. The contrastive models are trained on a Nvidia Titan V GPU for 4 epochs. The OSD model took 85 minutes to train and the UD model took 8 hours. We follow the training procedure from \citet{sun21} to estimate the kernel width parameters ($h_m$ in Equation~\ref{eqsemantic} and $h_{cf}$ in Equation~\ref{eqftsim}) via generative training when it is computationally feasible to do so and otherwise use 0.1 as our default value.

We check the similarity between two expressions in Equation~\ref{eqftsim} by comparing their fastText~\citep{bojanowski17} embeddings. For collaborative filtering, the neighborhood of words $L(S)$ in Equation~\ref{eqls} is defined as the 5 closest words (including the query word itself) in the dataset's slang expression vocabulary to the query word, measured in terms of cosine similarity between their respective fastText embeddings. We use the list of stopwords from NLTK~\citep{bird2009natural} to check whether a word is a content word. We apply the \textit{simple\_preprocess} routine from Gensim~\citep{rehurek11} before checking for the degree of content word overlap between two sentences.

\section{Additional Results}

\subsection{Additional Interpretation Examples} \label{appex}

Table~\ref{tableexample} show additional example interpretations made by the models evaluated in Section~\ref{evalmodel}. The first three examples illustrate cases where the semantically informed models were not able to predict the exact definitions, but came up with definitions that are more closely related to the groundtruth compared to the baseline. The latter two examples show cases where the semantically informed models fail to make an improvement.

\subsection{Effect of Context Length} \label{appcontext}

\begin{figure*}[t]
		\begin{center}
			\begin{subfigure}[b]{0.49\linewidth}
				\caption{OSD}
				\includegraphics[width=\linewidth]{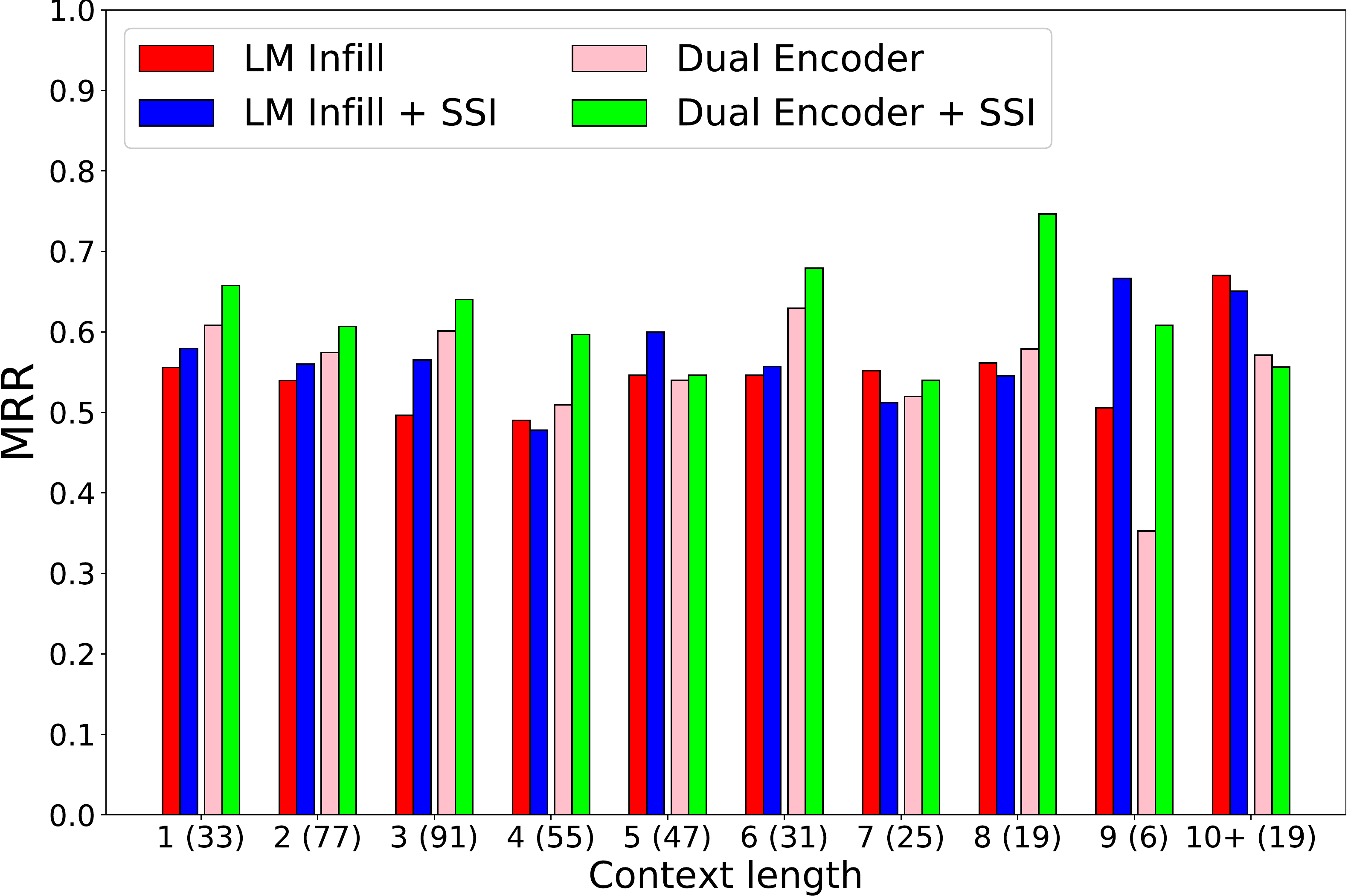}
			\end{subfigure} \hspace{0.05in}
			\begin{subfigure}[b]{0.49\linewidth}
				\caption{UD}
				\includegraphics[width=\linewidth]{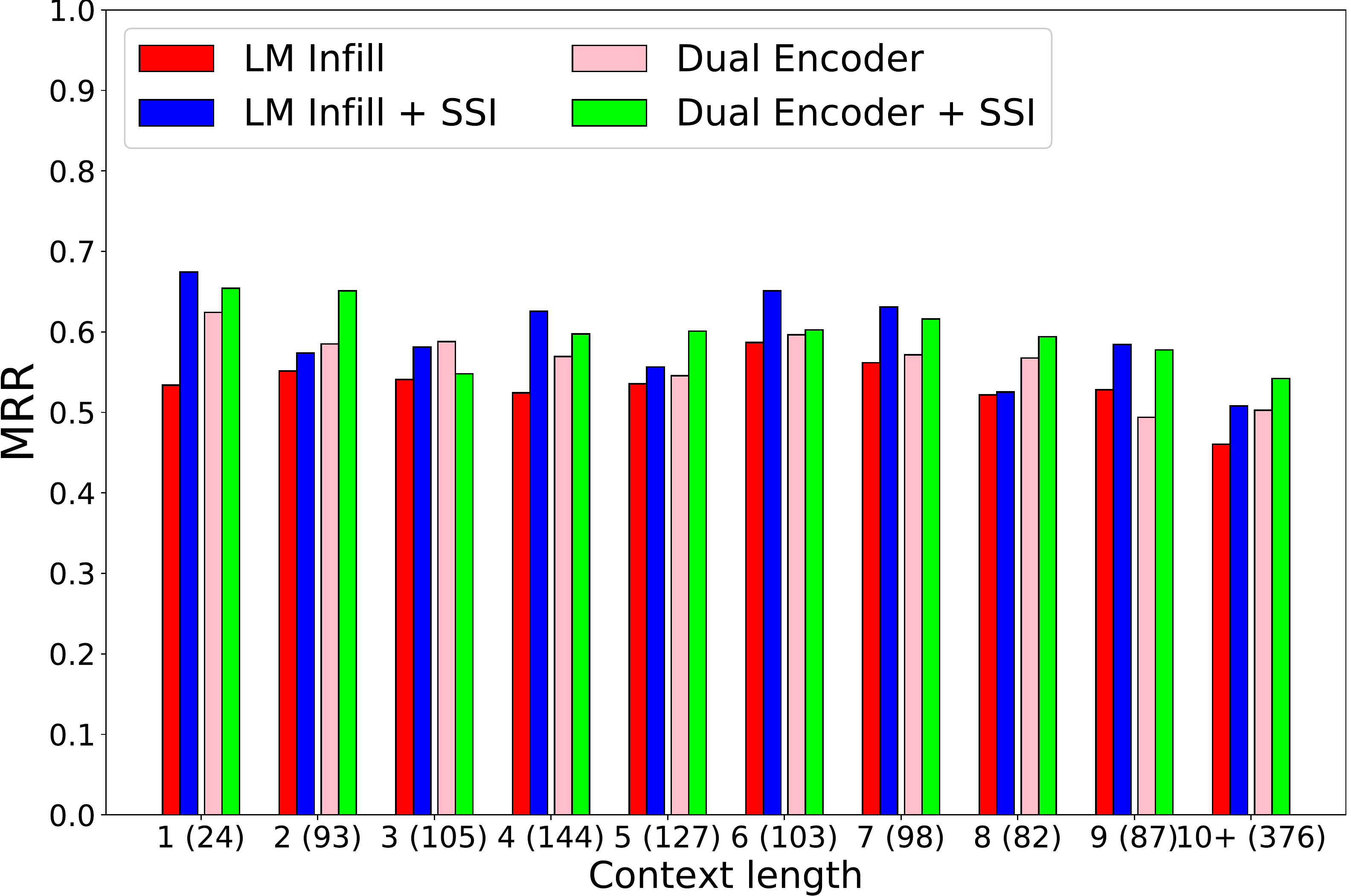}
			\end{subfigure}
		\end{center}
		\caption{Evaluation of slang interpretation performance measured in mean-reciprocal rank (MRR) for all models with $n$ = 50. Test entries are partitioned based on the number of content words (excluding the slang expression itself) found within the corresponding example usage sentence. Number of entries corresponding to each context length is shown in parenthesis on the x-axis legend.} 
		\label{figcontext}
\end{figure*}

In the model evaluation described in Section~\ref{evalmodel}, we control for the content-word length of the usage context sentence to examine its effect with respect to interpretation performance for both the baseline and the semantically informed models. Figure~\ref{figcontext} shows the results partitioned by the number of content words in the example usage sentence excluding the slang expression, evaluated against four distinctively sampled candidates. To our surprise, we do not observe any consistent trends when controlling for context length. Interpretation performance for both the context-based baseline models and their semantically informed variants is fairly consistent under different context length.

\subsection{Finetuning Dual Encoder} \label{appfinetune}

\begin{table}[t!]
	\centering\makebox[0.5\textwidth]{
		\begin{tabular}{lrr}
			Model&\makecell[r]{Distinct\\negatives}&\makecell[r]{Random  \\negatives}
			\\
			\addlinespace[0.05cm]
			\hline
			%\addlinespace[0.25cm]
			\addlinespace[0.1cm]
			Dual Encoder, $n$ = 5 &0.604&0.598\\
			%\addlinespace[0.1cm]
			\hspace{0.5cm} + SSI&\textbf{0.612}&\textbf{0.599}\\
			%\addlinespace[0.2cm]
			\addlinespace[0.1cm]
			Dual Encoder, $n$ = 50 &0.583&0.570\\
			%\addlinespace[0.1cm]
			\hspace{0.5cm} + SSI &\textbf{0.627}&\textbf{0.633}\\
	\end{tabular}}
	\caption{Interpretation results on OSD measured in mean-reciprocal rank (MRR) when training the Dual Encoder without filtering out entries corresponding to words in the OSD testset.}
	\label{tablefew2}
\end{table}

We consider the case of finetuning the Dual Encoder by training it on all available UD data entries and test on the full OSD test set. Under this scenario, the Dual Encoder model would have seen examples of slang in the OSD test set, though the difference between the definition sentences and usage examples would not allow it to memorize the exact answer. While examining how much knowledge can be transfered from one dataset to another, we also apply the SSI reranker trained on OSD training data on the finetuned results to simulate a stronger baseline model. Table~\ref{tablefew2} shows the results. When compared to the zero-shot results in Table~\ref{tableresults}, finetuning on entries corresponding to the same slang, albeit coming from two very different resources, does noticeably improve interpretation accuracy. Moreover, applying SSI to the improved interpretation candidates from the finetuned Dual Encoder further increases interpretation accuracy. 
This finding suggests that the improvement brought by SSI can indeed generalize in cases where the baseline context-based interpretation model outputs better interpretation candidates.

%This finding suggests that the performance of our framework is hindered by the baseline model's performance and the improvement can generalize in better context-based interpretation models.

\subsection{Machine Translation Examples} \label{apptrans}

Table~\ref{tableextrans} to Table~\ref{tableextrans4} show full example translations (English to French) made for the experiment described in Section~\ref{restrans}, translating sentences containing slang before and after applying slang interpretation.

\section{Data Permissions}

At the time when the research is performed, Online Slang Dictionary (OSD) explicitly forbids automated downloading of data from its website service. We therefore have obtained written permission from its owner to download and use the dataset for personal research use. We download data from the online version of the Oxford Dictionary (OD) under personal use. We cannot publically share the two datasets used above as a result. Readers interested in obtaining the exact datasets used in this work must first obtain relevant permission from the respective data owner before the authors of this work can share the data. The Urban Dictionary (UD) dataset is obtained from the authors of \citet{ni17} under a research only license. We release entries relevant to our study with the original data license attached.

\begin{table*}[t!]
	\centering\makebox[\textwidth]{
		\begin{tabular}{ll}
			\hline
			
            % Example 2
			\addlinespace[0.1cm]
			[Example 1]&\\
			\addlinespace[0.1cm]
			Query (target slang in \textit{\textbf{bold italic}}): &That girl has a \textit{\textbf{donkey}}.\\
			\addlinespace[0.1cm]
			Groundtruth definition of target slang: &Used to describe a girl's butt in a good way.\\
			\addlinespace[0.25cm]
			LM Infill baseline prediction: &Name, crush, boyfriend.\\
			\addlinespace[0.1cm]
			LM Infill + SSI prediction: &Horse, dog, puppy.\\
			\addlinespace[0.25cm]
			Dual Encoder baseline prediction: &Penis.\\
			\addlinespace[0.1cm]
			Dual Encoder + SSI prediction: &Girl with big ass and big boobs.\\
			\addlinespace[0.15cm]
            \hline
           
            % Example 3
			\addlinespace[0.1cm]
			[Example 2]&\\
			\addlinespace[0.1cm]
			Query: &I am an \textit{\textbf{onion}}.\\
			\addlinespace[0.1cm]
			Groundtruth definition of target slang: &A native of Bermuda.\\
			\addlinespace[0.25cm]
			LM Infill baseline prediction: &Adult, man, athlete.\\
			\addlinespace[0.1cm]
			LM Infill + SSI prediction: &Ren, adult, guard.\\
			\addlinespace[0.25cm]
			Dual Encoder baseline prediction: &An idiot.\\
			\addlinespace[0.1cm]
			Dual Encoder + SSI prediction: &An asian person.\\
			\addlinespace[0.15cm]
            \hline
            
            % Example 4
			\addlinespace[0.1cm]
			[Example 3]&\\
			\addlinespace[0.1cm]
			Query: &In Blastem version 4, they really \textit{\textbf{nerf}} the EnemyToaster.\\
			\addlinespace[0.1cm]
			Groundtruth definition of target slang: &In an update or sequel to a video game, to make a weapon weak or weaker,\\
			& such that it's like a Nerf gun.\\
			\addlinespace[0.25cm]
			LM Infill baseline prediction: &Were, called, attack.\\
			\addlinespace[0.1cm]
			LM Infill + SSI prediction: &Made, hacked, came.\\
			\addlinespace[0.25cm]
			Dual Encoder baseline prediction: &To do something.\\
			\addlinespace[0.1cm]
			Dual Encoder + SSI prediction: &To beat someone in the face with your penis.\\
			\addlinespace[0.15cm]
            \hline
            
            % Example 5
			\addlinespace[0.1cm]
			[Example 4]&\\
			\addlinespace[0.1cm]
			Query: &I heard Steve was sent to the \textit{\textbf{cooler}} for breaking and entering.\\
			\addlinespace[0.1cm]
			Groundtruth definition of target slang: &Reform school.\\
			\addlinespace[0.25cm]
			LM Infill baseline prediction: &School, house, class.\\
			\addlinespace[0.1cm]
			LM Infill + SSI prediction: &Bathroom, kitchen, grounds.\\
			\addlinespace[0.25cm]
			Dual Encoder baseline prediction: &Slang term for the police.\\
			\addlinespace[0.1cm]
			Dual Encoder + SSI prediction: &One of the most dangerous things in the world the best.\\
			\addlinespace[0.15cm]
            \hline
            
            % Example 6
			\addlinespace[0.1cm]
			[Example 5]&\\
			\addlinespace[0.1cm]
			Query: &Do you have any \textit{\textbf{safety}}\\
			\addlinespace[0.1cm]
			Groundtruth definition of target slang: &Marijuana.\\
			\addlinespace[0.25cm]
			LM Infill baseline prediction: &Money, friends, cash.\\
			\addlinespace[0.1cm]
			LM Infill + SSI prediction: &Self, shoes, money.\\
			\addlinespace[0.25cm]
			Dual Encoder baseline prediction: &Marijuana.\\
			\addlinespace[0.1cm]
			Dual Encoder + SSI prediction: &Word that is used to describe something that is very good.\\
			\addlinespace[0.15cm]
            \hline

	\end{tabular}}
	\caption{Additional examples: Example OSD slang entries with predicted definitions from both the language infill model (LM Infill) and the Dual Encoder model with $n$ = 50, along with predictions from the corresponding semantically informed slang interpretation (SSI) models.}
	\label{tableexample}
\end{table*}

\begin{table*}[t!]
	\centering\makebox[\textwidth]{
		\begin{tabular}{ll}
			\hline
			
			% Example 1
			\addlinespace[0.1cm]
			[Example 1]&\\
			\addlinespace[0.1cm]
			Query (target slang in \textit{\textbf{bold italic}}): &Let's smoke a \textit{\textbf{bowl}} of marijuana.\\
			Definition of target slang: &a marijuana smoking pipe.  Most frequently bowls are made\\ & out of blown glass, but can be made of metal, wood, etc.\\
			Groundtruth interpreted sentence: &Let's smoke a \textit{pipe} of marijuana.\\
			\addlinespace[0.25cm]
			Original query sentence translation: &Faisons fumer un bol de marijuana.\\
			&\small(\textit{BLEU:} 78.1, \textit{BLEURT:} 66.1, \textit{COMET:} 1.05)\\
			Gold-standard translation: &Faisons fumer une pipe de marijuana.\\
			\addlinespace[0.1cm]
			\hdashline
			\addlinespace[0.1cm]
			LM Infill interpretation \& translation:&\\
			\addlinespace[0.1cm]
			(1) Let's smoke a \textit{for} of marijuana. & Fumons un \textit{pour} de la marijuana.\\
			&\small(\textit{BLEU:} 47.1, \textit{BLEURT:} 20.6, \textit{COMET:} -0.58)\\
            (2) Let's smoke a \textit{in} of marijuana. & On fume un \textit{peu} (little) de marijuana.\\
            &\small(\textit{BLEU:} 51.6, \textit{BLEURT:} 64.8, \textit{COMET:} 0.48)\\
            (3) Let's smoke a \textit{myself} of marijuana. & Nous allons fumer \textit{moi-même} de la marijuana.\\
            &\small(\textit{BLEU:} 51.8, \textit{BLEURT:} 32.4, \textit{COMET:} -0.55)\\
            (4) Let's smoke a \textit{or} of marijuana. & Fumons un \textit{ou} de marijuana.\\
            &\small(\textit{BLEU:} 45.4, \textit{BLEURT:} 32.2, \textit{COMET:} -1.04)\\
            (5) Let's smoke a \textit{vapor} of marijuana. & Fumons une \textit{vapeur} de marijuana.\\
            &\small(\textit{BLEU:} 56.4, \textit{BLEURT:} 57.0, \textit{COMET:} 0.40)\\
            \addlinespace[0.25cm]
			LM Infill + SSI interpretation \& translation:&\\
			\addlinespace[0.1cm]
			(1) Let's smoke a \textit{pot} of marijuana. & Faisons fumer un \textit{pot} de marijuana.\\
			&\small(\textit{BLEU:} 79.5, \textit{BLEURT:} 78.8, \textit{COMET:} 1.15)\\
            (2) Let's smoke a \textit{pipe} of marijuana. & Faisons fumer une \textit{pipe} de marijuana. \\
            &\small(\textit{BLEU:} 100.0, \textit{BLEURT:} 99.1, \textit{COMET:} 1.32)\\
            (3) Let's smoke a \textit{pack} of marijuana. & Faisons fumer un \textit{paquet} de marijuana.\\
            &\small(\textit{BLEU:} 77.7, \textit{BLEURT:} 68.3, \textit{COMET:} 0.80)\\
            (4) Let's smoke a \textit{leaf} of marijuana. & Faisons fumer une \textit{feuille} de marijuana.\\
            &\small(\textit{BLEU:} 79.9, \textit{BLEURT:} 48.2, \textit{COMET:} 1.21)\\
            (5) Let's smoke a \textit{cigarette} of marijuana. & Faisons fumer une \textit{cigarette} de marijuana.\\
            &\small(\textit{BLEU:} 75.7, \textit{BLEURT:} 81.7, \textit{COMET:} 1.25)\\
			\addlinespace[0.15cm]
            \hline

	\end{tabular}}
	\caption{Additional examples of machine translation of slang, without or with the application of the SSI framework. The top 5 interpreted and translated sentences are shown for each model with BLEU, BLEURT, and COMET scores against the gold-standard translation shown in parentheses.}
	\label{tableextrans}
\end{table*}

\begin{table*}[t!]
	\centering\makebox[\textwidth]{
		\begin{tabular}{ll}
			\hline
			
			% Example 2
			\addlinespace[0.1cm]
			[Example 2]&\\
			\addlinespace[0.1cm]
			Query: &That band was so totally \textit{\textbf{vast}}.\\
			Definition of target slang: &Cool or anything good.\\
			Groundtruth interpreted sentence: &That band was so totally \textit{cool}.\\
			\addlinespace[0.25cm]
			Original query sentence translation: &Ce groupe était si vaste.\\
			&\small(\textit{BLEU:} 53.2, \textit{BLEURT:} 32.9, \textit{COMET:} -0.59)\\
			Gold-standard translation: &Ce groupe était tellement cool.\\
			\addlinespace[0.1cm]
			\hdashline
			\addlinespace[0.1cm]
			LM Infill interpretation \& translation:&\\
			\addlinespace[0.1cm]
			(1) That band was so totally \textit{popular}. & Ce groupe était tellement \textit{populaire}.\\
			&\small(\textit{BLEU:} 74.5, \textit{BLEURT:} 78.7, \textit{COMET:} 0.43)\\
            (2) That band was so totally \textit{good}. & Ce groupe était si \textit{bon}.\\
            &\small(\textit{BLEU:} 51.8, \textit{BLEURT:} 77.0, \textit{COMET:} 0.32)\\
            (3) That band was so totally \textit{different}. & Ce groupe était complètement \textit{différent}.\\
            &\small(\textit{BLEU:} 57.2, \textit{BLEURT:} 50.3, \textit{COMET:} -0.07)\\
            (4) That band was so totally \textit{famous}. & Ce groupe était si \textit{célèbre}.\\
            &\small(\textit{BLEU:} 54.4, \textit{BLEURT:} 66.2, \textit{COMET:} -0.21)\\
            (5) That band was so totally \textit{new}. & Ce groupe était totalement \textit{nouveau}.\\
            &\small(\textit{BLEU:} 64.2, \textit{BLEURT:} 50.2, \textit{COMET:} -0.21)\\
            \addlinespace[0.25cm]
			LM Infill + SSI interpretation \& translation:&\\
			\addlinespace[0.1cm]
			(1) That band was so totally \textit{huge}. & Ce groupe était tellement \textit{énorme}.\\
			&\small(\textit{BLEU:} 81.1, \textit{BLEURT:} 56.0, \textit{COMET:} 0.15)\\
            (2) That band was so totally \textit{big}. & Ce groupe était tellement \textit{grand}.\\
            &\small(\textit{BLEU:} 83.0, \textit{BLEURT:} 50.7, \textit{COMET:} -0.19)\\
            (3) That band was so totally \textit{important}. & Ce groupe était si \textit{important}.\\
            &\small(\textit{BLEU:} 55.9, \textit{BLEURT:} 49.9, \textit{COMET:} -0.58)\\
            (4) That band was so totally \textit{cool}. & Ce groupe était tellement \textit{cool}.\\
            &\small(\textit{BLEU:} 100.0, \textit{BLEURT:} 97.9, \textit{COMET:} 1.29)\\
            (5) That band was so totally \textit{bad}. & Ce groupe était si \textit{mauvais}.\\
            &\small(\textit{BLEU:} 52.3, \textit{BLEURT:} 62.9, \textit{COMET:} -0.48)\\
			\addlinespace[0.15cm]
            \hline

	\end{tabular}}
	\caption{Continuation of Table~\ref{tableextrans}.}
	\label{tableextrans2}
\end{table*}

\begin{table*}[t!]
	\centering\makebox[\textwidth]{
		\begin{tabular}{ll}
			\hline
			
			% Example 3
			\addlinespace[0.1cm]
			[Example 3]&\\
			\addlinespace[0.1cm]
			Query (target slang in \textit{\textbf{bold italic}}): &Man, I ain't been to that place in a \textit{\textbf{fortnight}}!\\
			Definition of target slang: &An unspecific, but long-ish length of time.\\
			Groundtruth interpreted sentence: &Man, I ain't been to that place in a \textit{long time}!\\
			\addlinespace[0.25cm]
			Original query sentence translation: &Je ne suis pas allé à cet endroit en une quinzaine!\\
			&\small(\textit{BLEU:} 36.1, \textit{BLEURT:} 61.2, \textit{COMET:} 0.57)\\
			Gold-standard translation: &Je n'y suis pas allé depuis longtemps!\\
			\addlinespace[0.1cm]
			\hdashline
			\addlinespace[0.1cm]
			LM Infill interpretation \& translation:&\\
			\addlinespace[0.1cm]
			(1) Man, I ain't been to that place in a \textit{while}! & Je ne suis pas allé à cet endroit depuis un \textit{moment}!\\
			&\small(\textit{BLEU:} 46.9, \textit{BLEURT:} 76.5, \textit{COMET:} 0.88)\\
            (2) Man, I ain't been to that place in a \textit{million}! & Je ne suis pas allé à cet endroit dans un \textit{million}!\\
            &\small(\textit{BLEU:} 38.8, \textit{BLEURT:} 25.1, \textit{COMET:} -1.17)\\
            (3) Man, I ain't been to that place in a \textit{both}! & Je ne suis pas allé à cet endroit dans les \textit{deux}!\\
            &\small(\textit{BLEU:} 42.2, \textit{BLEURT:} 25.7, \textit{COMET:} -0.98)\\
            (4) Man, I ain't been to that place in a \textit{vanilla}! & Mec, je n'ai pas été à cet endroit dans une \textit{vanille}!\\
            &\small(\textit{BLEU:} 16.2, \textit{BLEURT:} 7.3, \textit{COMET:} 1.53)\\
            (5) Man, I ain't been to that place in a \textit{ignment}! & Mec, je n'ai pas été à cet endroit dans un \textit{ignement}!\\
            &\small(\textit{BLEU:} 16.2, \textit{BLEURT:} 12.7, \textit{COMET:} -1.31)\\
            \addlinespace[0.25cm]
			LM Infill + SSI interpretation \& translation:&\\
			\addlinespace[0.1cm]
			(1) Man, I ain't been to that place in a \textit{week}! & Je ne suis pas allé à cet endroit en une \textit{semaine}!\\
			&\small(\textit{BLEU:} 38.2, \textit{BLEURT:} 49.8, \textit{COMET:} 0.45)\\
            (2) Man, I ain't been to that place in a \textit{minute}! & Je ne suis pas allé à cet endroit en une \textit{minute}!\\
            &\small(\textit{BLEU:} 38.8, \textit{BLEURT:} 42.5, \textit{COMET:} -0.36)\\
            (3) Man, I ain't been to that place in a \textit{hour}! & Je ne suis pas allé à cet endroit en une \textit{heure}!\\
            &\small(\textit{BLEU:} 38.7, \textit{BLEURT:} 35.8, \textit{COMET:} -0.51)\\
            (4) Man, I ain't been to that place in a \textit{decade}! & Je n'y suis pas allé depuis une \textit{décennie}\\
            &\small(\textit{BLEU:} 68.8, \textit{BLEURT:} 81.8, \textit{COMET:} 1.03)\\
            (5) Man, I ain't been to that place in a \textit{day}! & Je ne suis pas allé à cet endroit en une \textit{journée}!\\
            &\small(\textit{BLEU:} 37.1, \textit{BLEURT:} 49.7, \textit{COMET:} -0.30)\\
            \addlinespace[0.15cm]
            \hline

	\end{tabular}}
	\caption{Continuation of Table~\ref{tableextrans2}.}
	\label{tableextrans3}
\end{table*}

\begin{table*}[t!]
	\centering\makebox[\textwidth]{
		\begin{tabular}{ll}
			\hline
            
            % Example 4
			\addlinespace[0.1cm]
			[Example 4]&\\
			\addlinespace[0.1cm]
			Query: &I want to go get coffee but it's \textit{\textbf{bitter}} outside.\\
			Definition of target slang: &Abbreviated form of bitterly cold.\\
			Groundtruth interpreted sentence: &I want to go get coffee but it's \textit{bitterly cold} outside.\\
			\addlinespace[0.25cm]
			Original query sentence translation: &Je veux aller prendre un café mais c'est amer dehors.\\
			&\small(\textit{BLEU:} 65.0, \textit{BLEURT:} 59.8, \textit{COMET:} 0.77)\\
			Gold-standard translation: &Je veux aller prendre un café, mais il fait très froid dehors.\\
			\addlinespace[0.1cm]
			\hdashline
			\addlinespace[0.1cm]
			LM Infill interpretation \& translation:&\\
			\addlinespace[0.1cm]
			(1) I want to go get coffee but it's \textit{raining} outside. & Je veux aller prendre un café mais il \textit{pleut} dehors.\\
			&\small(\textit{BLEU:} 68.1, \textit{BLEURT:} 79.9, \textit{COMET:} 0.97)\\
            (2) I want to go get coffee but it's \textit{closed} outside. & Je veux aller prendre un café mais il est \textit{fermé} dehors.\\
            &\small(\textit{BLEU:} 70.7, \textit{BLEURT:} 53.9, \textit{COMET:} -0.15)\\
            (3) I want to go get coffee but it's \textit{pouring} outside.& Je veux aller chercher du café, mais ça \textit{coule} dehors.\\
            &\small(\textit{BLEU:} 51.9, \textit{BLEURT:} 31.6, \textit{COMET:} -0.38)\\
            (4) I want to go get coffee but it's \textit{been} outside. & Je veux aller prendre un café, mais ça a \textit{été} dehors.\\
            &\small(\textit{BLEU:} 68.4, \textit{BLEURT:} 27.1, \textit{COMET:} -0.88)\\
            (5) I want to go get coffee but it's \textit{starting} outside & Je veux aller prendre un café, mais ça \textit{commence} dehors.\\
            &\small(\textit{BLEU:} 68.5, \textit{BLEURT:} 31.0, \textit{COMET:} -0.57)\\
            \addlinespace[0.25cm]
			LM Infill + SSI interpretation \& translation:&\\
			\addlinespace[0.1cm]
			(1) I want to go get coffee but it's \textit{cold} outside. & Je veux aller prendre un café, mais il fait \textit{froid} dehors.\\
			&\small(\textit{BLEU:} 90.3, \textit{BLEURT:} 92.7, \textit{COMET:} 1.20)\\
            (2) I want to go get coffee but it's \textit{warm} outside. & Je veux aller prendre un café mais il fait \textit{chaud} dehors.\\
            &\small(\textit{BLEU:} 78.1, \textit{BLEURT:} 79.1, \textit{COMET:} 1.12)\\
            (3) I want to go get coffee but it's \textit{driving} outside. & Je veux aller prendre un café mais il \textit{conduit} dehors.\\
            &\small(\textit{BLEU:} 70.4, \textit{BLEURT:} 26.5, \textit{COMET:} -0.69)\\
            (4) I want to go get coffee but it's \textit{closing} outside. & Je veux aller prendre un café mais il se \textit{ferme} dehors.\\
            &\small(\textit{BLEU:} 69.8, \textit{BLEURT:} 23.2, \textit{COMET:} -0.81)\\
            (5) I want to go get coffee but it's \textit{dark} outside. & Je veux aller prendre un café, mais il fait \textit{noir} dehors.\\
            &\small(\textit{BLEU:} 82.3, \textit{BLEURT:} 73.7, \textit{COMET:} 0.80)\\

			\addlinespace[0.15cm]
            \hline

	\end{tabular}}
	\caption{Continuation of Table~\ref{tableextrans3}.}
	\label{tableextrans4}
\end{table*}

\end{document}